# SEMPAI: a Self-Enhancing Multi-Photon Artificial Intelligence for prior-informed assessment of muscle function and pathology

Alexander Mühlberg[1]*, Paul Ritter[1,2], Simon Langer[3], Chloë Goossens[4], Stefanie Nübler[1], Dominik Schneidereit[1,2], Oliver Taubmann[3], Felix Denzinger[3], Dominik Nörenberg[5], Michael Haug[1], Wolfgang H. Goldmann[6], Andreas K. Maier[3], Oliver Friedrich[1,2], Lucas Kreiss[1,2]

[1] Institute of Medical Biotechnology, Department of Chemical and Biological Engineering,
  Friedrich-Alexander University Erlangen-Nuremberg, Germany

[2] Erlangen Graduate School in Advanced Optical Technologies, Germany

[3] Pattern Recognition Lab, Department of Computer Science,
  Friedrich-Alexander University Erlangen-Nuremberg, Germany

[4] Clinical Division and Laboratory of Intensive Care Medicine,
  KU Leuven, Belgium

[5] Department of Radiology and Nuclear Medicine, University Medical Center Mannheim,
  Medical Faculty Mannheim, Germany

[6] Biophysics Group, Department of Physics,
  Friedrich-Alexander University Erlangen-Nuremberg, Germany

\* Corresponding Author: alexander.mueale.muehlberg@fau.de

**Abstract**

Deep learning (DL) shows notable success in biomedical studies. However, most DL algorithms work as a black box, exclude biomedical experts, and need extensive data. We introduce the Self-Enhancing Multi-Photon Artificial Intelligence (SEMPAI), that integrates hypothesis-driven priors in a data-driven DL approach for research on multiphoton microscopy (MPM) of muscle fibers. SEMPAI utilizes meta-learning to optimize prior integration, data representation, and neural network architecture simultaneously. This allows hypothesis testing and provides interpretable feedback about the origin of biological information in MPM images. SEMPAI performs joint learning of several tasks to enable prediction for small datasets.

The method is applied on an extensive multi-study dataset resulting in the largest joint analysis of pathologies and function for single muscle fibers. SEMPAI outperforms state-of-the-art biomarkers in six of seven predictive tasks, including those with scarce data. SEMPAI's DL models with integrated priors are superior to those without priors and to prior-only machine learning approaches.

# 1. Introduction

Muscle tissue forms a hierarchically structured biological system and its morphology is closely linked to its function[1-3]. The parallel alignment of sarcomeres within a single myofibril and of myofibrils within a fiber results in parallel force vectors for effective force generation. A variety of muscle pathologies affect this structured muscle morphology, leading to reduced function of the entire system. Duchenne muscular dystrophy (DMD), for instance, is caused by an overall loss of structural integrity in individual fibers, eventually leading to failure of respiratory and heart muscle that can be life-limiting[4]. Besides chronic degenerative diseases like DMD, also acute myopathies can result in disruptions of the myofibrillar structural alignment, as it has been shown in ongoing sepsis[5].

Muscle morphology can be revealed by modern macroscopic and microscopic imaging modalities, such as magnetic resonance imaging (MRI)[6], computed tomography (CT)[7], ultrasound[8], and second harmonic generation (SHG)[9] in multi-photon microscopy (MPM). All these techniques are considered label-free, as they do not require exogenous markers or extensive tissue preparation. Thereby, artefacts such as swelling, shrinkage, or altered mechanical properties are limited. However, only SHG imaging provides sub-µm resolution to resolve sarcomeres (~2µm in size), which are relevant to establish a deeper understanding of the structure-function relationship and the impact of pathologies on single muscle fibers[10, 11].

Function of muscle tissue is based on its passive mechanical and its active force generation properties. Passive force parameters are related to the visco-elastic behavior of the muscle. In contrast, active force parameters describe its intrinsic ability to generate force, e.g., represented by the physiological sensitivity to calcium ions. Automated integrated biomechatronics systems, such as the *MyoRobot*[3, 12] or the *MechaMorph* system[13], can measure these active and passive parameters simultaneously alongside the imaging process. Both aforementioned systems consist of force transducers (FT) to measure force and voice coil actuators (VCA) to perform axial movement with higher precision as compared to stepper motors[14]. A combination of high-resolution label-free SHG microscopy with biomechanical measurements of active and passive force was recently demonstrated[13]. Through this, correlations between morphological features derived from SHG and functional properties acquired with FTs and VCAs were experimentally shown for individual muscle fibers from *mdx* and wild type (WT) mice.

Research into automated assessment of muscle pathology and function from MPM image data using artificial intelligence (AI), specifically machine learning (ML) and neural networks (NN), seems a promising approach to advance in the understanding of muscle structure-function relationships. However, for the assessment of muscle fibers using label-free SHG microscopy, the use of AI has not yet been demonstrated at all.

The field of optical microscopy in general has benefited from a broad variety of ML applications[15], such as automation[16, 17], segmentation[18], and image quality (IQ) enhancement including optimal illumination[19], emitter localization in super-resolution microscopy[20], or image restoration[21]. Modern ML approaches for microscopy also start to integrate prior knowledge of imaging physics. For instance, the integration of physics knowledge into the learning process of an AI helped with the technological optimization of microscope- and software-components[22, 23] for enhanced IQ, and with digital staining of virtual fluorescence in label-free phase microscopy[24], or Fourier ptychography microscopy[25]. An integration of physics simulations even allowed 3D modelling of sub-resolution viruses in diffraction-limited microscopy[26] or the subcellular image segmentation in data sets from different microscopes[27].

An emerging field of AI research is meta-learning, or "learning to learn", that analyzes which conditions must be given to be able to effectively learn a specific task. This includes the relatively new field of neural architecture search (NAS)[28], with the goal to automatically identify a suitable NN architecture for a given problem. This might replace the time-consuming trial-and-error process of manual architecture search and may not only provide competitive performance, but also solutions with particularly desireable properties, such as curiosity[29]. On the downside, NAS, and more generally meta-learning, are computationally expensive approaches, although a variety of techniques are developed to decrease time and associated costs[28]. Recently, a novel meta-learning approach for segmentation problems in biomedical imaging gained a lot of attention: nnU-Net[30]. nnU-Net optimizes NN architecture and hyperparameters together with rule-based image processing operations (normalization, resampling etc.), with the eponymous U-Net serving as the base NN architecture. This approach outperformed most prevailing methods for a large number of automated segmentation problems in biology and medicine[30].

Although DL has shown its strengths for big data, as in the case of automated classification of images in the world wide web, for fundamental medical research with limited data sets, methods based on prior knowledge can show competitive performance for describing or predicting a pathology. Providing prior biological knowledge, or in brief "priors", to the learning algorithm as a baseline instead of starting from scratch, therefore, seems plausible. Another common drawback of many DL systems is the lack of explainability. A large number of methods, such as DeepSHAP[31], are developed to highlight the image information relevant for the decision-making process. However, a fundamental question posed by Rudin[32] was why the current research focuses on post-hoc explanations of complicated models rather than creating more interpretable models from the beginning. Explainability can be increased by using priors, such as established measurements, in the learning process of a NN. Additionally, the integration of prior knowledge in the form of known operators as NN layers was already shown to stabilize the learning process by reducing the maximum error bounds[33]. Lastly, by integration of priors, human understanding and intuition about a problem can be employed within AI research.

Based on these current trends, we present a novel AI with a broader objective than mere prediction: the Self-Enhancing Multi-Photon Artifical Intelligence (SEMPAI) is specifically designed to integrate hypothesis-driven priors in a data-driven deep learning (DL) approach for research on MPM images of muscle fibers. SEMPAI as a general tool simultaneously identifies optimal data representation, degree of prior integration, and NN architecture for a given biomedical problem. In contrast to the technologically-inspired optimization of microscope parameters for enhanced IQ, it performs biologically-inspired meta-learning on already existing databases. Additionally, it leverages common patterns shared over all prediction tasks to enable the learning with small data sets. SEMPAI therefore aims to integrate the hypotheses of researchers and identify biologically relevant information with scarce data.

We apply SEMPAI to an extensive single muscle fiber multi-study data collection, consisting of high quality image data acquired with label-free multiphoton imaging systems and highly-automated function assessment by robotized biomechatronics, to predict and understand a variety of functional and pathological muscular properties.

## 2. Results
## 2.1. Selected studies, functional parameters, pathologies, and priors

We retrospectively screened in-house experimental MPM data acquired over more than a decade to obtain a large database that includes a variety of biological properties with respect to muscle pathology and function. We name these properties *labels*. In each of four studies (A-D), murine muscle fibers were imaged using the same multiphoton microscope. From these images, morphological image features were computed with previously reported software[34]. In brief, these features include the *cosine angle sum* (CAS) taken from selected 2D planes (2D-CAS) and in 3D (3D-CAS), the *vernier density* (VD), the 3D sarcomere length (3D-SL), and the cross-sectional area (CSA) of single fibers. Since these features have already been

## a. SHG Imaging and Force Recordings of Single Muscle Fibres

**Single Fiber Preparation**

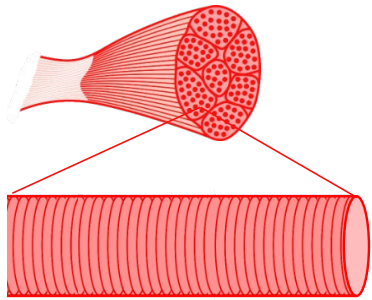

Assignment of Physiological Labels
- Healthy WT
- Dystrophic Phenotype *mdx*
- Inflammatory Phenotype *Sepsis*
- Muscle Type

**Label-free SHG Microscopy**

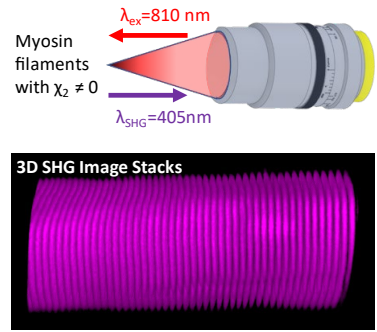

Myosin filaments with $\chi_2 \neq 0$
$\lambda_{ex}$=810 nm
$\lambda_{SHG}$=405nm

3D SHG Image Stacks

Structure Assessment via Priors
- Cosine Angle Sum (CAS) in 2D and in 3D
- Vernier Density (VD)
- 3D Sarcomere length (3D-SL)

**Robotized Biomechatronics**

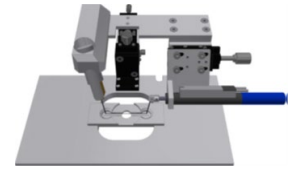

Force Recordings

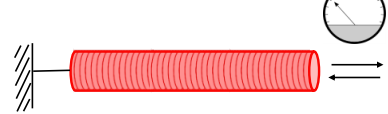

Function Assessment
- Active: $Ca^{2+}$-Activated Force Transients Curve
- Passive: Visco-Elasticity Curve

## b. Automated Standardization of Data Sets

**Automated Cross-Study Standardization of Images**

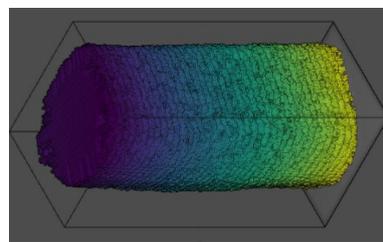

Computation of Fiber Cross-Sectional Area (CSA)

Isotropic Resampling → Denoising & Background Removal → Contrast Enhancement → Multi-Scale Rigid Registration → Cropping to Probable Location Bounding Box

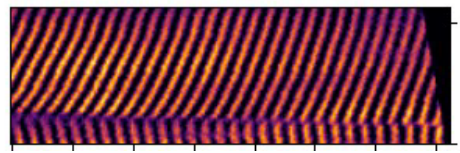

**Automated Extraction of Function Labels from Force Curves**

Active Force
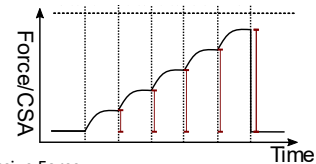

Passive Force
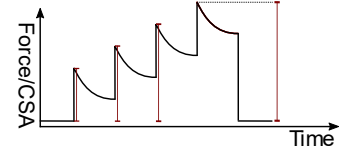

Sensitivity to Calcium Signaling
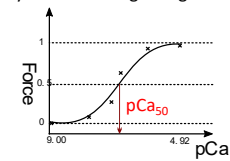
$pCa_{50}$

## c. Distribution of Labels and Priors in Standardized Data

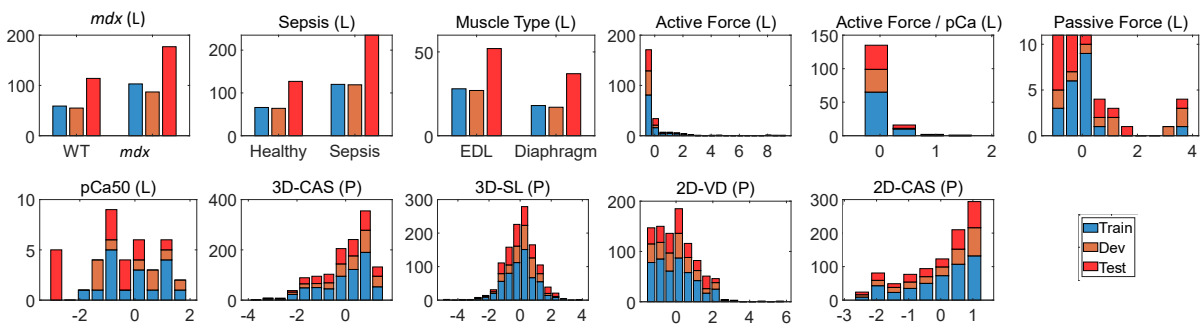

**Figure 1** Data acquisition, cross-study standardization, and value distribution of labels and priors in train, dev, and test set. **a.** Single muscle fibers were dissected from murine muscle tissue. The data were annotated regarding pathologies and muscle type. In each case, 3D label-free second harmonic generation (SHG) microscopy was performed, and morphological features, termed priors, were calculated. Muscle tissue was assessed for its function by robot-assisted biomechanical force measurements. **b.** The SHG images are standardized with a dedicated image processing pipeline consisting of resampling, denoising, registration, contrast enhancement and cropping of the images to a probable location bounding box. Within this standardization process, the cross-sectional area (CSA) of the fibers is calculated. Function labels like active force or pCa50 are automatically computed from the raw curves coming from the biomechatronics system. **c.** Distribution of priors (P) and labels (L) in train, development (dev) and test data after stratified grouped data split. The distributions are normalized to standard score.

**Table 1.** Used multi-study data after standardization and exclusion of data with inappropriate IQ. Please note that the total number of unique images is **not** a sum of all above, since most images had information for labels, e.g., an image from mdx, where active force was available. Example images for each of the included studies are shown in Supporting Information 1. WT: wild type, C: classification, R: regression.

| Label / Task | Data set with reference | Total number of curated images |
| --- | --- | --- |
| Inflammatory phenotype: Sepsis/WT, C | A[1] | 731 |
| Dystrophic phenotype: *mdx*/WT, C | B1[2], B2[2], C[2], D[3] | 567 |
| Muscle type: Diaphragm/EDL, C | D[3] | 179 |
| Active Force, R | B1[2] | 232 |
| Active Force/pCa, R | B1[2] | 152 |
| Passive (Restoration) Force, R | C[2] | 39 |
| pCa50, R | B2[2] | 39 |
| **Total number of unique images** | | **1,298** |

shown to be descriptive for a variety of rather specific remodeling patterns in muscle research, related to aging, chronic degenerative or inflammatory myopathies[5, 13, 35, 36], we use them as prior information, and term them accordingly as *priors*. A more elaborate explanation of the extraction of priors is given in the **Methods.**

**Table 1** shows the investigated learning tasks, the corresponding original studies, and the number of samples used. The extended variants, i.e., larger sample size, of the following studies are included in our database: (A) For investigating muscle atrophy during sepsis, samples from the *extensor digitorum longus* (EDL) of septic and WT mice were imaged and complemented by active force recordings in EDL single fibers of the same animal[5]. (B1) Active force measurement and subsequent SHG imaging at each force recording was carried out in EDL single fibers from WT and *mdx* mice[13]. (B2) Force recordings from a different image data set of EDL fibers to deduce the $Ca^{2+}$ sensitivity of the contractile apparatus, pCa50, as a measure for the troponin-C $Ca^{2+}$ sensor characteristics[3, 37]. (C) The same setting was used to access passive force parameters on a different set of animals[13]. (D) Fixated single fibers and fiber bundles from EDL and diaphragm in *mdx* and WT animals were imaged to investigate structural differences between *mdx* and WT as well as between the muscle types of EDL and diaphragm.

## 2.2 Cross-study data standardization

The workflow for data acquisition, standardization, and the resulting data distribution is shown in **Figure 1**. Standardization is required to compare images from different studies acquired under varying experimental conditions and during different time periods. By standardization, the technical variance can be minimized. Further fine-grained details of the implementation are provided in the Methods.

In brief, the images are resampled to an isotropic voxel size of 0.5μm, slightly denoised via a median filter, and the background, which is defined by Otsu's thresholding of the image, is set to zero. Then, the Multidimensional Contrast Limited Adaptive Histogram Equalization (MCLAHE) algorithm[38] is applied for contrast enhancement of each muscle fiber. This contrast-enhanced image is registered to a pre-selected fiber with canonical orientation and fiber pattern by a rigid multi-scale registration, and the resulting transformation is applied to the non-enhanced version. A mean image of the registered fibers is created after setting all foreground voxels to one, which provides the probability of presence for muscle fibers in each voxel. The bounding box of voxels with probability >0.85 is generated (extent: 180x80x57μm³) and applied to the images. This cropping of images to relevant regions enables the use of DL models with reduced degrees of freedom (DOF), which is advantageous for our data regime. For a standardized computation of the CSA, we have developed an algorithm that integrates an outlier detection with three different methods, namely i) simple counting of muscle fiber pixels within each slice after thresholding and re-orientation, ii) a principal component analysis (PCA)-based and iii) an ellipsoid envelope fit-based computation of the CSA (for more details see Methods section). Automatically generated reports of the standardization were used by two independent researchers with domain knowledge to exclude images with obviously non-acceptable quality from the extended database, resulting in N=1,298 unique 3D images used for our analysis. However, images with low signal-to-noise ratio but visible muscle fiber were not excluded. Examples for the IQ of standardized images of each study with/without contrast enhancement are shown in **Supporting Information 1**.

The standardized data are stratified and grouped into training (train), development (dev) and test set (2/4, 1/4, 1/4). The grouping prevents different images of single fibers extracted from the same muscle bundle from being distributed over different sets, which would result in information leakage. The stratification ensures that the distribution in the respective sets is similar, thus, label instances with rare occurrence are present in the train, dev and test sets. The label and prior data are normalized to a standard score based on mean and standard deviation of the train set.

## 2.3 SEMPAI method overview

During its self-enhancement process, SEMPAI simultaneously optimizes configurations of its three main components: the prior integration, the data representation (DaRe), and NN architecture and hyperparameters (NN settings).

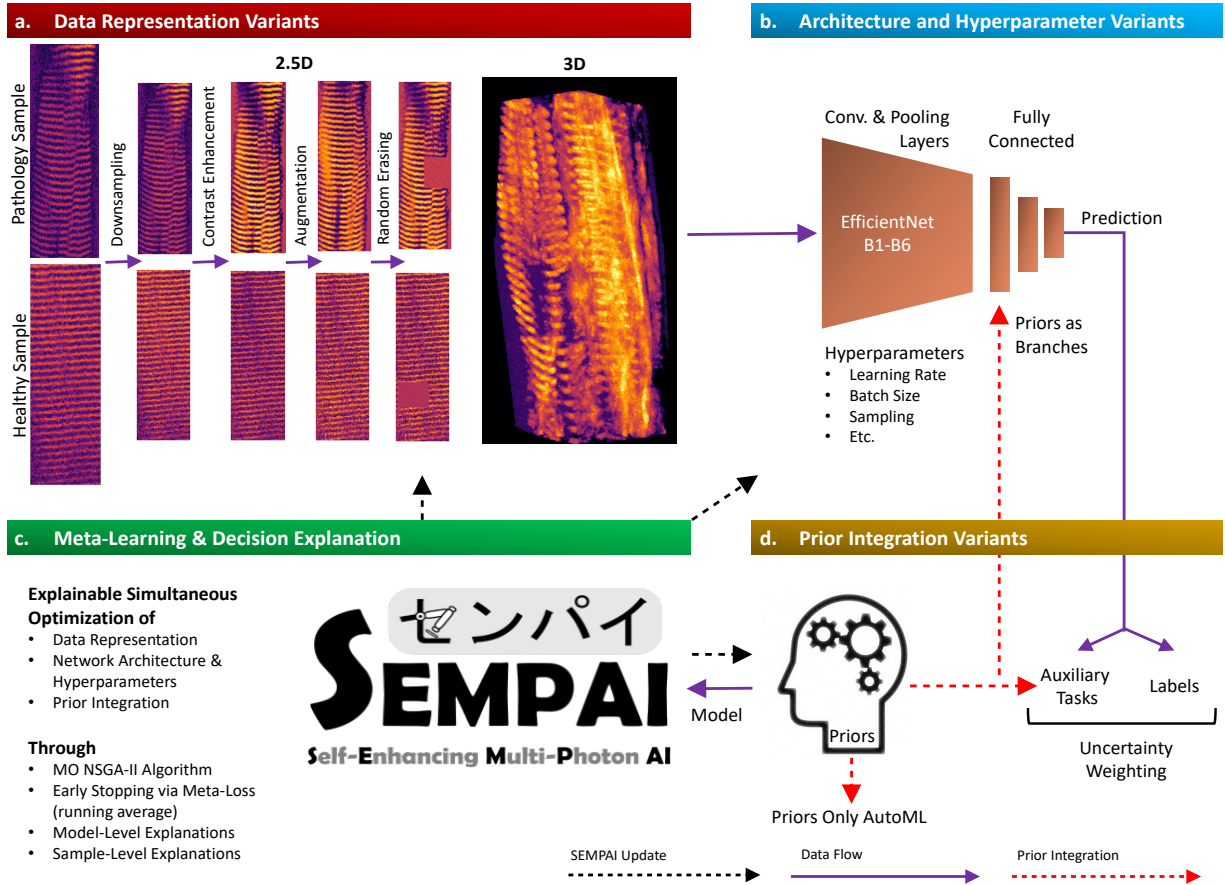

**Figure 2. SEMPAI method overview.** For each iteration, termed trial, of-the-self-enhancement process, a data representation (DaRe) is selected that represents the images either in 2.5D, i.e., three regularly spaced slices are used, or in 3D. Then, decisions are made regarding the modification of the DaRe such as down-sampling or contrast enhancement. The selected DaRes are fed to a NN, and the NN architecture and its hyper-parameters are selected. The level of prior integration is then chosen. SEMPAI decides, whether priors (i) are not employed, (ii) are used as auxiliary tasks for the NN training, (iii) are fed directly to the fully connected layer of the NN as branches, or (iv) are used in both integration methods, i.e., a combination of (ii) and (iii). In option (v), the priors are the only input to an AutoML approach for handcrafted features. The resulting model of the trial is used to predict the labels on the dev set. The performance of the model yields the meta-losses that guide SEMPAI's configuration selection for the next trial. This process results in a simultaneous self-enhancement of DaRe, NN architecture&hyperparameters and prior integration with increasing number of trials. Table 2 shows the configuration space.

The method is shown in **Figure 2**, the configuration space in **Table 2**, an extended rationale as well as details about implementation in the Methods.

SEMPAI can choose from five different levels to integrate priors (or hypotheses). It can learn without priors (*NoPriors*), use them as auxiliary tasks (*AuxLosses*), which results in a soft constraint to the learning problem[39], integrate them as *Branches* into the fully connected layer of the NN or a combination of both (*AuxLosses&Branches*). In the fifth configuration (*PriorsOnly*), only priors are used in an integrated AutoML method[40] for handcrafted features, i.e., without using the SHG images and DL. To the best of our knowledge, the integration within SEMPAI is the first attempt to combine current priors (biomarkers) known in single fiber muscle research with ML.

**Table 2.** Configuration space. Decisions made by SEMPAI during self- enhancement process.

| Data Representation Variants | | | | |
|---|---|---|---|---|
| **Contrast Enhancement** | *Yes*: the MCLAHE algorithm is applied on the images | *No*: No further enhancement after registration and resampling | | |
| **Down-sampling** | *Yes*: Images are down-sampled to 0.75µm voxel size isotropically | *No*: No resampling, 0.5µm isotropically | | |
| **Augmentation** | *Yes*: Application of 3D augmentation such as Gaussian noise, rotation, flipping, affine transformation | *No*: Original standardized images are used | | |
| **Random Erasing** | *Yes*: Random regions of the image are erased | *No*: Original standardized images are used | | |
| **Volume/slice selection** | *3D*: The whole 3D array of each sample is used | *2.5D_1*  *2.5D_5*  *2.5D_10*  *2.5D_20* Center slice and 2 slices with 1, 5, 10 or 20µm distance to the center slice are selected | | |
| **Prior Integration Variants** | | | | |
| **Method** | *NoPriors*   *AuxLosses* | *Branches* | *AuxLosses& Branches* | *PriorsOnly* |
| **NN Settings Variants** | | | | |
| **Capacity** | 2D/3D EfficientNet *B1-B6* | | | |
| **Learning rate** | Cyclic (*Yes/No*) and in range [0.0001, 0.2] | | | |
| **Optimizer** | Adam or SGD with Nesterov moment | | | |
| **Momentum** | Momentum in range [0.9, 0.99] | | | |
| **Gradient Clipping** | *Yes:* gradients are clipped to the norm 1.0 | *No:* NN gradients evolve freely | | |
| **Batch Size** | 2.5D/3D: *small* (32/4); *medium* (64/8), *large* (128/16), *XL* [256/32], *XXL* (512/64) | | | |
| **Imbalance Sampling** | *Yes:* class distributions are re-balanced based on strata information of the initial train-dev-test split. Sampling weights are estimated automatically | *No*: The original data distribution is fed in the NN | | |

The decisions by SEMPAI regarding DaRe indicate "how and where" the biological information can optimally be learned. For example, SEMPAI analyzes whether *3D* images are needed or whether three regularly spaced representative slices (*2.5D*) are sufficient and how large this spacing should be. Analogously, SEMPAI provides information on the importance of down-sampling, which can help to estimate the required image resolution for a learning task. As a side effect, such feedback may also have an impact on future studies as larger pixel sizes result in shorter scan times, increasing experimental throughput.

For NN training, in addition to hyperparameter optimization, SEMPAI selects one architecture variant from the base architecture EfficientNet[41] that offers variants with different capacity (*B1* to *B6*) and for 2D or 3D, and has been shown to yield competitively predictive performance with less DOF than alternative architectures[41]. SEMPAI learns all tasks jointly in a multi-task setup: Our hypothesis is that this enforces a semantic regularization of the learning process, since systematic differences unrelated to the biological origin, e.g., in IQ, are less likely to be used for prediction. Instead, the use of related muscle-specific patterns across different learning tasks is enforced. Joint learning further has the advantage that tasks with small data can still be learned, as DOF are determined by information from similar tasks[42]. Recent research shows that joint learning is preferable to the similar concept of transfer learning[43]. In case of missing labels for either primary or auxiliary losses of a sample, no backpropagation occurs during training for the corresponding model outputs, i.e., these outputs are "masked" for that sample. This results in a sort of interleaved learning, in which different tasks are learned in different batches. It also enables joint training without the need for data imputation. During NN training, all losses are weighed against each other by uncertainty weighting[44].

The resulting model of each trial created on the train set is applied for prediction of the labels on the dev set. The predictive performance of the chosen model for each task is assessed for the dev set. Those performances are used as meta-losses to select the configurations for the next trial. SEMPAI uses NSGA-II[45] multi-objective optimization for this selection, i.e., there is not only one loss to be minimized, but the losses of all labels are minimized independently.

For tasks with small data, i.e., pCa50 and passive force (cp. Table 1), ML and especially DL are severely limited due to overfitting. Based on the identified associations of the same priors with the investigated labels in previous studies[5, 13, 35, 36], we hypothesize that our muscle-specific learning tasks are related and the mean predictive performance over all tasks may assist SEMPAI to select a regularized model. Therefore, a *total meta-loss* is introduced, which is a weighted sum of all meta-losses for each task and provides an estimate of the model performance over all tasks. This loss is not used for optimization, but for selection of the models for small data tasks.

SEMPAI explains itself regarding (i) decision-making during the self-enhancement process (SEMPAI model-level explanations) as well as regarding (ii) the decision-relevant image pixels/voxels and priors of each sample (SEMPAI sample-level explanations): (i) For each task, based on the performed experiments and their results, SEMPAI retrospectively fits a random forest model to estimate its predictive performance from a given configuration. Subsequently, the fitted model is fed in the SHAP Tree Explainer[46] to estimate the impact of DaRe, NN settings, and prior integration and identify configurations that yield models with good predictive performance. (ii) For the sample-level explanation of important image regions, SEMPAI utilizes DeepSHAP[31]. In the case of prior integration method *Branches* or combined *AuxLosses&Branches*, the method was extended to provide attribution of priors together with, and orthogonally to, the attribution map of the image.

## 2.4 Predictive performance of SEMPAI and comparison with benchmarks

The predictive performance is given for the unseen test set (holdout) as area under the curve (*AUC*) of the receiver operating characteristic for classification tasks and as $R^2$ for regression tasks.

To benchmark the performance of SEMPAI, we implemented two state-of-the-art (SOTA) baselines: As univariate analyses still reflect the standard approach in biomarker research, we select the best prior on the combined train and dev set and use it as a univariate predictor for the test set. In addition, to assess the performance of SOTA multivariate modelling, we use all priors and fit a statistical pipeline, consisting of MRMR[47] feature selection, best subset selection and multiple linear/logistic regression, on the combined train and dev set, and apply the model for the prediction on the test set. For fair benchmarking, as statistical models are more severely regularized, potentially resulting in underfitting, we vary the best subset selection information criterion (Akaike/Bayesian) and the penalty of the regression (L2/elastic net: L1&L2) and report the best performance on the test set. To understand the merit of priors and images individually, we report the results of SEMPAI when using only priors (SEMPAI *PriorsOnly*), i.e., when it does not have access to the images, and the opposite, i.e., exclude trials that integrated priors (SEMPAI *NoPriors*). Finally, to test susceptibility of SEMPAI for non-optimal configurations, we give the average performance of the 50 best models (SEMPAI50).

The results of SEMPAI, in detail and with train and dev set performance, and the comparison with SOTA are shown in **Figure 3**. In six of seven investigated learning tasks, SEMPAI was superior to SOTA models in predicting the labels of the test set.

For prediction of dystrophic phenotype *mdx* and the muscle type, SEMPAI achieved an *AUC* of 0.93 for both while SOTA achieved 0.78 and 0.80, respectively, using the prior 2D-VD in both cases. Predictive performance for the inflammatory phenotype sepsis was generally lower, and SEMPAI was only on par with SOTA with prior 3D-SL, with *AUC* 0.77 in both cases.

Active force was predicted by SEMPAI with $R^2$ 0.37, while SOTA gave 0.20 using the prior 2D-CAS. The prediction of the biologically more interesting active force adjusted for pCa yielded similar results with a performance of $R^2$ 0.39 by SEMPAI and 0.21 for SOTA by prior 3D-CAS. For passive force, SEMPAI again achieved solid results with $R^2$ 0.33, while SOTA achieved 0.23 via the multivariate model. For pCa50, SEMPAI was only slightly superior, $R^2$ 0.24, to using the prior 3D-CAS, $R^2$ 0.19.

Predictions of force parameters were more susceptible to performance decrease for non-optimal configurations than those of pathologies and muscle type, evident from the results for SEMPAI50, which in the case of the force predictions showed inferior results compared to the best trial.

**Table 3.** SEMPAI results overall in train, dev and test set; of SEMPAI sub-configurations, and comparison with SOTA methods (all results on the test set if not denoted otherwise). ns: negative sign, i.e., worse than guessing.

| Task | SEMPAI Train/Dev/Test | SEMPAI NoPriors/PriorsOnly | SEMPAI50 | SOTA Multiv. Model | SOTA Best Prior |
|---|---|---|---|---|---|
| mdx [AUC] | 1.0/0.96/0.93 | 0.93/0.87 | 0.92 | 0.70 | 2D-VD: 0.78 |
| Sepsis [AUC] | 0.94/0.82/0.77 | 0.68/0.75 | 0.74 | 0.77 | 3D-SL: 0.77 |
| Muscle Type [AUC] | 1.0/0.95/0.93 | 0.93/0.86 | 0.88 | 0.67 | 2D-VD: 0.80 |
| Active Force [$R^2$] | 0.82/0.66/0.37 | 0.14/0.31 | 0.13 | 0.03 | 2D-CAS: 0.20 |
| Active Force/pCa [$R^2$] | 0.97/0.67/0.39 | 0.06/0.35 | 0.19 | 0.04 | 3D-CAS: 0.21 |
| Passive Force [$R^2$] | 0.91/0.74/0.33 | ns/0.08 | 0.16 | 0.23 | 2D-CAS: 0.20 |
| pCa50 [$R^2$] | 0.45/0.07/0.24 | ns/ns | ns | 0.01 | 3D-CAS: 0.19 |

As expected, the prediction for tasks of smaller sample size, pCa50 and passive force, was problematic for models with large DOF or without strict regularization as shown by the predictive performance of DL (SEMPAI *NoPriors)*, single-task AutoML (SEMPAI *PriorsOnly*) and, in case of pCa50, even a simple multivariate statistical model with only few DOF. A regularization by joint learning, integration of priors, and the model selection based on the *total meta-loss*, however, resulted in a SEMPAI model with slightly improved performance compared to the best SOTA approach, the univariate predictor 3D-SL (one DOF).

In three of seven tasks, SEMPAI *PriorsOnly* was superior to SEMPAI *NoPriors* and especially achieved competitive performance in classification tasks and for predicting active force. The priors already provided the diagnostic information for classifying the inflammatory phenotype sepsis, since no improvement in predictive performance was observed by additional utilization of DL on images.

In contrast, for the dystrophic phenotype *mdx* and the muscle type, SEMPAI *NoPriors* yielded very strong models and, in the case of *mdx*, these predictions were superior to those based solely on priors. Thus, the performance of *PriorsOnly* or *NoPriors* models varied largely between tasks. In all tasks, however, SEMPAI identified a level of prior integration on the dev set that led to a good generalizability, i.e., the best predictive performance for the test set.

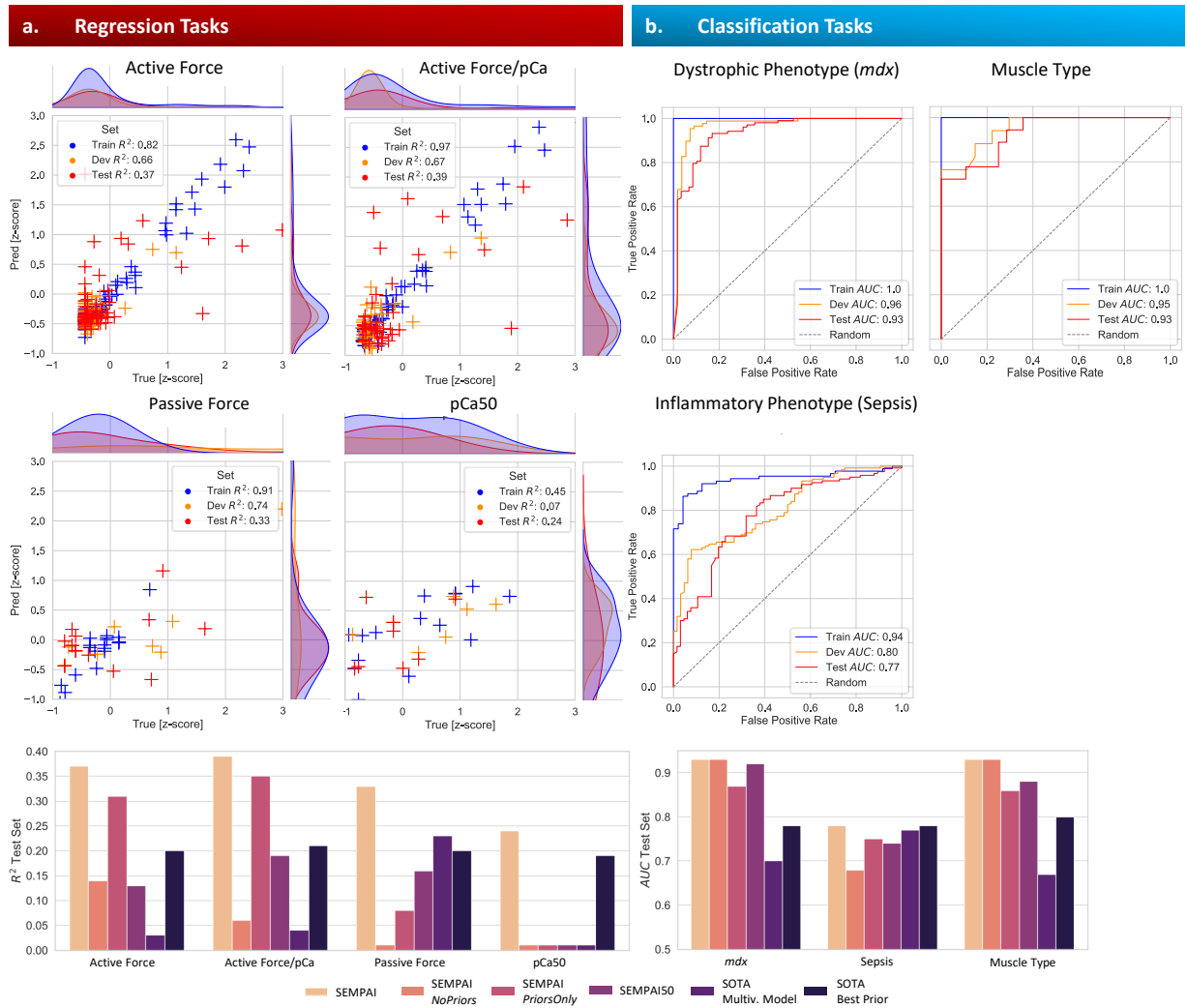

**Figure 3.** SEMPAI overall results of SEMPAI, its sub-configurations and comparison with state-of-the-art (SOTA) methods. Performance metrics in train (for NN training), dev (for meta-optimization) and test set (unseen data) for regression (**a.**), $R^2$, and classification, $AUC$, tasks (**b.**).

Especially for the prediction of muscle function, synergistic effects of combining prior knowledge with DL are observed, as SEMPAI provided strongly improved performance compared to DL without priors or models solely based on priors. These effects may be interpreted as a DL-based prior (or hypothesis) refinement.

## 2.5 SEMPAI as a tool to obtain relevant biological information and localize it

As described above, to explain its decision-making on model-level, SEMPAI computed the respective SHAP values of the samples and the mean absolute SHAP values over all samples to quantify the association of the configuration space with the predictive performance. In addition, the stability of the analysis was tested (see section Methods). The results are shown in **Figure 4**.

In five of seven investigated tasks, the level of prior integration was the most important decision. For the classification tasks *mdx*, sepsis, and muscle type, integration of priors was especially important according to the mean absolute SHAP values. While *mdx* and muscle type preferred the soft constraint of priors as *AuxLosses*, the harder learning task of predicting sepsis preferred a stronger integration of priors as *AuxLosses&Branches*. This is plausible since the univariate predictor 3D-SL was also on par with SEMPAI for predicting sepsis (cp. Figure 3). Although selecting the level of prior integration was on average not the most important decision for learning muscle function, the highest positive impact on predictive performance was found with strong prior integration, namely *Branches* and *AuxLosses&Branches*, for active and passive force. The results for pCa50 are harder to interpret. According to the individual SHAP values, the task preferred no prior integration or weak integration as *AuxLosses* but the pattern of the association is rather complex.

Most learning tasks, especially *mdx* and sepsis, benefited from smaller NN capacity, indicating that fewer DOF were sufficient for the complexity of the task and helped to avoid overfitting.

Reducing image resolution had a negative impact on five of seven learning tasks, although at varying degrees, as indicated by the SHAP values when employing down-sampling. Especially *mdx* profited from a higher resolution.

Prediction of active and passive force benefited from contrast enhancement. This is also intuitive when inspecting the images visually, as the IQ for function assessment is lower on average due to a more complex experimental setup[13] (Supporting Information 1). On the contrary, the modification of image intensities by contrast enhancement had a negative effect for the tasks *mdx*, muscle type and sepsis. This indicates that not only the structure but also the original intensity yields important information for these tasks and should not be artificially modified.

In two of seven tasks, the selection of the spacing between three representative slices was the most important decision. Interestingly, for active and passive force prediction, SEMPAI strongly profited from using slices from the periphery of the muscle fiber (±20μm), compared to using further slices in the proximity of the muscle center (i.e., 1, 5 and 10μm). This indicates additional biological information for function assessment in the muscle periphery in comparison to a sole evaluation of the muscle center. Using configurations that employed DaRe 3D generally provided an inferior predictive performance, and none of these models was found among the top-50 models for any task.

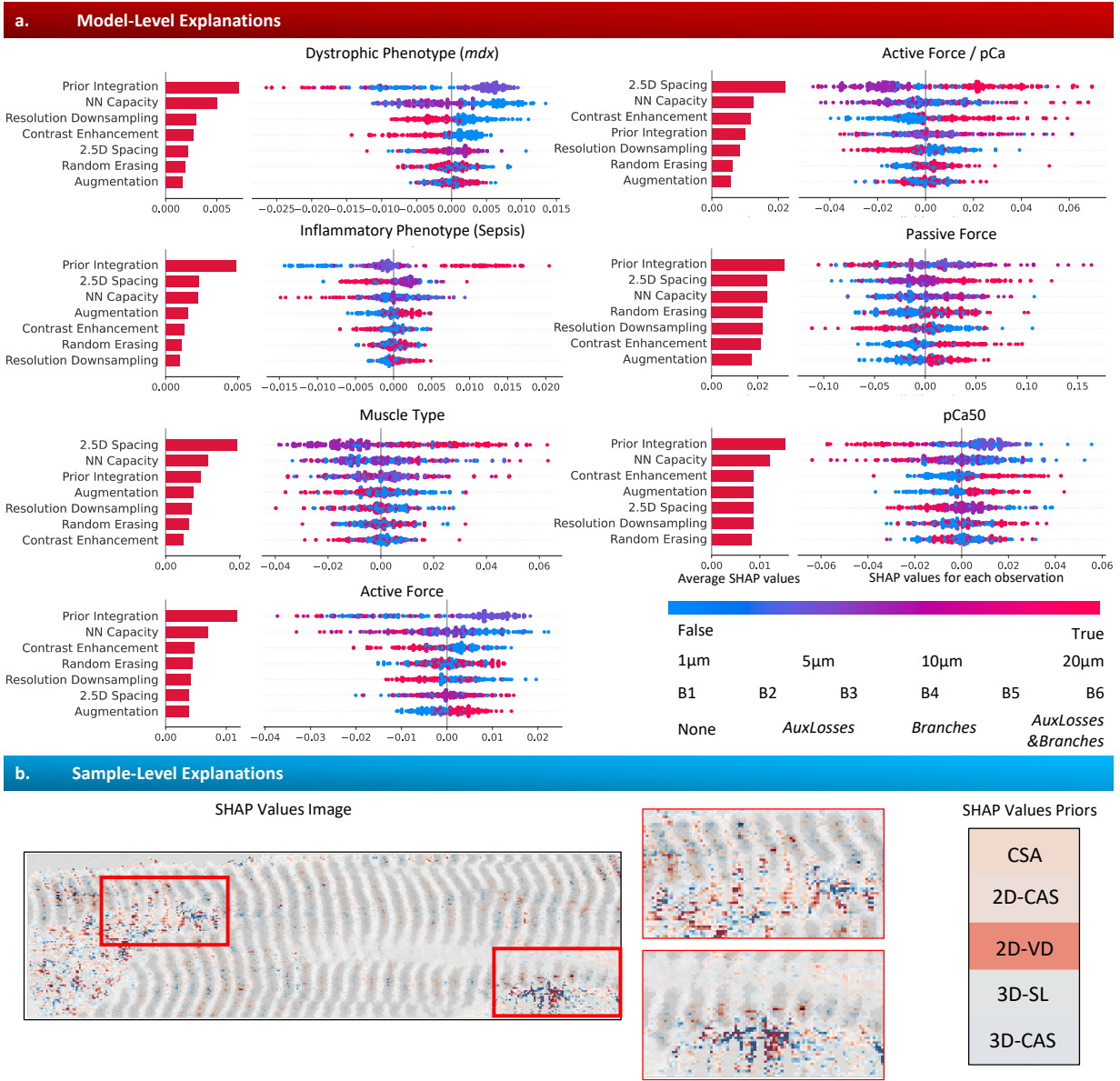

**Figure 4** Decision visualization regarding the self-enhancement process, i.e., model-level explanations (a.), and regarding decision-relevant image voxels/pixels and priors, i.e., sample level-explanations (b.). **a.** A random forest model learns the predictive performance of SEMPAI for a specific label as a function of the configuration space. The resulting model is then analyzed by SHAP Tree Explainer which allows to estimate the individual contribution of each configuration for each sample in units of the performance metrics ($AUC/R^2$). Decisions are sorted top-to-bottom based on their mean absolute SHAP values as a surrogate for the importance of the decision. Configurations are color-coded from weak to strong expression of a configuration (legend in lower right). **b.** Attribution map of image (left) and priors (right) for one *mdx* sample. Colored voxels and priors are used by SEMPAI for this sample to correctly predict *mdx*. The attribution of priors is computed simultaneously and shown with the same color code and scale.

*mdx* was the only task for which localized properties were of special interest, as the performance decreased by employing random erasing[48]. To investigate this effect, we used the sample-level decision explanation of SEMPAI. This confirmed that in addition to 2D-VD, localized regions of twisted or damaged muscle fibers were especially used to predict *mdx* (Figure 4b.). When those image regions were randomly erased, a loss of predictive performance was observed. For all other tasks, however, more global properties seem to be important for prediction, making the augmentation effect[48] of random erasing more advantageous.

As seen exemplarily for *mdx* (Figure 4b), SEMPAI provides a detailed case-wise highlighting of important image regions orthogonally to the information given by priors. A collection of examples is shown in **Supporting Information 2**. However, for a proper quantitative evaluation, those observations have to be validated in a standardized manner, which is beyond the scope of this study. In the future, an observer study based on SEMPAI could lead to novel scientific insights.

## 3 Discussion

We developed a novel Self-Enhancing Multi-Photon Artificial Intelligence (SEMPAI) and applied it on a total of 1,298 single muscle fiber 3D MPM images. SEMPAI was superior to previous state-of-the-art (SOTA) biomarkers in predicting active and passive muscle force, pCa50 for $Ca^{2+}$-activated isometric force, muscular dystrophy phenotype in the *mdx* mouse as well as muscle type. To the best of our knowledge, deep learning (DL) was not yet applied to MPM image data bases in single muscle fiber research. DL was already applied to gene data from DMD patients[49] or to perform functional evaluation of DMD on ultrasound images[50]. In this context, DL is most often used on MRI databases, e.g., for the identification of MRI biomarkers in smaller cohorts (N=26)[51], for image classification[52] or for the analysis of larger clinical cohorts (N=432)[53]. However, ultrasound and MRI do not offer sufficient resolution to understand DMD and the *mdx* model at the level of individual muscle fibers. Here, MPM has the unique advantages of label-free image contrast and sub-cellular resolution.

SEMPAI targets close interaction with the biomedical researcher, on the one hand by integrating, testing and refining prior knowledge or hypotheses of the domain expert and on the other hand by giving systematic feedback about influencing factors for optimal extraction of biologically relevant information. The researcher can therefore use his domain knowledge as input to the method and receives comprehensible and easy-to-interpret feedback as output.

Most studies with DL develop their neural network (NN) architecture for a fixed data representation (DaRe). SEMPAI uses the simultaneous optimization of the DaRe for biological knowledge discovery. For instance, we showed that most of the investigated learning tasks, as expected, benefit from a higher image resolution. SEMPAI further showed that the muscle periphery is especially important for the assessment of active and passive force measurements or that the distinctive properties of *mdx* dystrophic phenotype are rather learned locally, i.e., at specific locations of the fiber, than globally, i.e., widespread over the whole fiber. However, prediction of *mdx* by SEMPAI is to a certain extent also possible using only global characteristics, which is in concordance with recent literature[54]. The information provided by SEMPAI can be used to guide future experiments and to refine microscopy hardware specifically for a pathology, e.g., by maintaining high resolution in the case of *mdx* or by decreasing resolution in the case of sepsis to increase throughput. While the recent ground-breaking meta-learning approach of Isensee et al. to the biomedical image segmentation problem[30] is more technically driven by evolving its decision-making around pre-processing and network topology, SEMPAI focuses its decision-making rather on integrating and returning interpretable information regarding prior knowledge and biology.

Usually, imaging-based biomarker studies are either purely based on priors or novel DL architectures. Our study reveals that a prior integration, by varying degrees, in DL methods almost always yielded the best predictive performance, especially for the prediction of muscle function. Recent research, such as known-operator learning[33], points in a similar direction and has already shown impressive results by integrating known operators, e.g., subtasks with known analytic solutions in image reconstruction algorithms, into NNs to improve task performance, while preserving the reliability of deterministic methods[33]. However, the decision to integrate priors in known operator learning is a design choice made before the experiments are conducted. SEMPAI's approach is agnostic and decides based on the current task, if priors are needed. The regularization by weak constraints in the form of auxiliary losses[39] is particularly interesting as this variant of regularization, in addition to competitive predictive performance for our data, has the benefit of being able to process samples, in which priors are not available or not reliable due to low IQ. SEMPAI has learned the priors during training and implicitly uses them for inference of those cases even without explicit prior computation. A similar concept of regularization, but for dynamical systems, is applied in physics-informed neural networks[55], which regularize the learning of systems dynamics by known differential equations. Priors are represented by the differential equations that are incorporated into the NN training by losses that use the deviation between predictions made by the NN and those expected following the equations.

SEMPAI leverages shared patterns using joint learning. The benefit of jointly learning multiple tasks has been shown previously[42, 56]; it allows for a more robust prediction performance even in those tasks for which only a few positive samples are available. Otherwise, with just a small number of examples insufficient for training a high-variance model from scratch, relying on an already established biomarker would often be the only option. Notably, joint learning is also interesting for biological reasons, as shown in pan-cancer research[57], since the highlighting of common patterns might be beneficial in the development of appropriate drugs. In addition to joint learning of multiple tasks by the NN, it was suggested that joint meta-learning, i.e., simultaneous optimization of NN architecture and configurations over different tasks might be beneficial[29]. This is explicitly utilized by SEMPAI as well.

As a key limitation of this study, while intended as a general-purpose tool, SEMPAI was only evaluated for muscle research. In the future, we plan to apply the method to gastroenterological and pneumonological MPM data bases and priors. Further, SEMPAI did not yield a good predictive performance with 3D DL based on the underlying architecture. The phenomenon that DL approaches using lower-dimensional "multi-view" data representations are sometimes superior to DL methods working directly on 3D data is well-known[58, 59]. Further conceptual developments are required for beneficial use of full 3D information.

# 4 Methods

*Selected studies:*
   a. *A – inflammatory phenotype (sepsis vs. control)[5].* Sepsis was induced by caecal ligation and puncture (CLP) of 24-week-old mice and the *extensor digitorum longus* (EDL) muscle was extracted. Maximum isometric tetanic forces were induced in the native whole muscle via needle electrodes (Aurora Scientific) by averaging three consecutive tetanic stimuli (150Hz stimulation frequency, 200ms duration, 0.2ms pulse width, 2 min rest intervals). Thereafter, the dissected and in paraformaldehyd (PFA) fixed muscle tissue was imaged with a voxel size of 0.2×0.2×0.5μm, in a field-of-view of 100×100 μm with a stack depth of typically 50μm. Single fiber biomechanics was assessed using the previously described *MyoRobot* system to measure active force and reconstruct the force-pCa curve. The 3D-SL and myofiber diameter were derived at the beginning of the experiment.
   b. *B1 & B2 – active force & dystrophic phenotype (mdx vs. WT)[13].* The age of the mice was between 13 and 21 weeks for WT and between 27 and 91 weeks for *mdx*. Single muscle fiber segments were manually dissected from the native EDL muscle and clamped into the *MechaMorph* system for subsequent force measurements and SHG imaging. The fiber was adjusted so that its SL was in the range of 2.2 – 3.1μm as shown by the *MechaMorph* system. Then, force measurements were performed to assess active force parameters (see above). The maximum activation was measured at a pCa of 4.92 in an undiluted highly activating solution (HA, mM: Hepes 30, Mg(OH)$_2$ 6.05, EGTA 30, CaCO$_3$ 29, Na$_2$ATP 8, Na$_2$CP 10, pH 7.2). Specific force, Hill-fit and pCa50 were determined. SHG imaging was performed in two different scenarios (B1 & B2). In B1, a 3D SHG image stack was recorded at each single force recording. In B2, the fiber was only imaged in the relaxed state (pCa 9). Single fibers were z-scanned using a 0.5 μm step size and at a voxel-size of $0.139 \times 0.139 \times 0.500$ μm$^3$.
   c. *C - passive force & dystrophic phenotype (mdx vs. WT)[13].* The overall procedure was the same as in the active force measurements described above (see B1&B2). However, in this case the *MechaMorph* system was used to access passive force parameters. At each step of force recording, an SHG 3D image stack of the fiber was recorded before proceeding to the next stretch step.
   d. *D – muscle type (EDL vs. SOL) & dystrophic phenotype (mdx vs WT) in tissue.* The investigated mice were 9 months of age. Whole muscle tissue from EDL and diaphragm was fixated in 4% PFA and transferred in PBS on dry-ice during transportation. Each muscle was cut longitudinally at the highest cross-sectional area. Small cryo-cuts of 10μm were performed and collected on microscope slides. Each slice was further investigated by SHG microscopy. VD, CAS and SL were derived. In some cases (N=222), images were recorded from whole muscle tissues. In these cases, single fibers were digitally cropped from the 3D image stacks and afterwards standardized. Force recordings were not performed.

*Label-free SHG imaging and functional force measurements:*
   a. *Label-free SHG imaging.* Label-free SHG imaging was performed on an inverse multiphoton microscope (TriMScope, LaVision BioTec, Bielefeld, Germany) with a mode-locked fs-pulsed Ti:Sa laser (Chameleon Vision II, Coherent, Santa Clara, CA, USA). The laser was tuned to a wavelength of 810nm, generating the second harmonic generation signal at 405nm. The laser was focused into the sample by a water immersion objective (LD C-Apochromat lens - 40x/1.1/UV-VIS-IR/WD 0.62, Carl Zeiss, Jena, Germany), and the generated SHG signal was detected by an ultra-sensitive photo multiplier tube (PMT) (H 7422-40 LV 5M, Hamamatsu Photonics) in transmission mode to preliminary target the SHG of myosin-II.

b. *Functional force measurements via the MyoRobot system*[3, 12]. The MyoRobot is a biomechatronics system for automated assessment of biomechanical active and passive properties as previously described.
   c. *Functional force measurements via the MechaMorph system*[13]. The *MechaMorph* is a custom-engineered device for combined structure–force measurements. A small measurement chamber can be inserted onto the microscope stage below the objective. Single muscle fiber segments can be mounted between a force transducer and a software-controlled voice coil actuator (VCA) that allows the *MechaMorph* to perform subsequent isometric force measurements and structural imaging via SHG microscopy.

*Priors:*
   a. *Cosine angle sum (CAS).* The CAS quantifies the angular deviation of myofibrillar bundles from the main axis [34]. This well-established parameter was deduced from 2D planes of SHG images by a software algorithm (2D-CAS) [34, 60, 61]. The CAS describes disturbances in muscle myofibrillar architecture that have been shown to correlate with muscle weakness[35]. Recently, an upgraded version for 3D assessment of CAS (3D-CAS) was developed[5].
   b. *Vernier density (VD).* Y-shaped deviations from the regular sarcomere pattern in z-stacks of SHG images are referred to as "verniers". The number of these verniers is then normalized to the fiber area to obtain the VD. Values close to zero represent fibers, where all myofibrils are perfectly in register, while larger values indicate deviations. The VD can either be generated manually or by a custom-designed software tool[61].
   c. *Sarcomere length (SL).* The software tools for *MechaMorph* and *MyoRobot*, the SL can be displayed live, while the new 3D_CAS software[5] records the SL as additional image parameter.
   d. *Smart Cross-Sectional Area (CSA) computation.* In the current study, we report a new method for quantifying the CSA of single muscle fibers, which has been developed for a standardized solution of the CSA in all image data sets. First, a binarization of the images is performed by a simple Otsu threshold on the images. An oriented bounding box algorithm[62] is applied to the binarized fiber to orient the fiber vertically. The top and bottom 10 slices are excluded from quantification. Then, three algorithms are combined with each other, and an outlier detection is applied to increase the stability of the method.
      i. Algorithm 1 - exact counting: Since the binarized fiber is now arranged from top to bottom, morphological operations 2D opening and closing are applied to each slice to close holes and obtain a compact segmentation. After application, the number of pixels in each slice is counted and averaged.
      ii. Algorithm 2 - principal component-based: Instead of morphological operations, a 2D principal component analysis (PCA) of scikit-learn is applied and the obtained maximum and minimum radii were used to determine the area of an ellipse for each slice. The results are averaged over the slices.
      iii. Algorithm 3 - elliptic envelope-based: Instead of morphological operations, an elliptic envelope (EE) is calculated with a contamination of 0.2. The area of the EE is calculated for each slice and the results are averaged across slices.

   The mean results of two algorithms, which show higher concordance, are used. The averaging and outlier removal compensates for potential weaknesses of the algorithms due to varying IQ. The results agreed well with visual assessment.

*Implementation of cross-study standardization and data split:* The pipeline was written in Python (v3.7.7). For studies with low SNR, a median filter of size 1μm was applied. We define an intensity threshold for the background by Otsu's thresholding. Then, voxels with intensities below this threshold intensity were set to 0 (background). The contrast enhancement algorithm MCLAHE[38] was applied with adaptive histogram range. The registration toolbox Elastix[63] was used to register the muscle fibers to a reference fiber, which exhibits a canonical structure and perfectly vertical orientation. We used a rigid multi-scale Euler registration with 600 iterations, automatic scale estimation, center of gravity initialization, 32 bins, 6 scales, and grid-adaptive step size. The transformation was then also applied to the non-enhanced fiber. Each standardized fiber is normalized to a sample-wise standard score. Force measurements are extracted directly from the TDMS curves coming from the instruments, entered into the data frame and normalized by the CSA of the associated fiber. The standardization pipeline is highly automated, and the steps are documented by an automatically generated SEMPAI labbook to identify and minimize errors associated with standardization or data management.

For data splitting in train (2/4), dev (1/4) and test (1/4) set, the data are both stratified and grouped. The stratification is needed to have sufficient data with a certain label in all sets. Continuous functional labels are median-dichotomized into "high" and "low" values, e.g., specific force "high" for stratification. However, those dichotomized labels are only used for stratification and not as a learning task. This stratification also ensures that class distributions are balanced over train, dev and test set. The labels are grouped according to muscle bundle, single fibers from one bundle are therefore, not split between train, dev and test set, preventing information leakage.

*Implementation of SEMPAI configuration-space and self-enhancement* SEMPAI was implemented in Python (v3.8.1). NN parts in PyTorch (v1.11, CUDA v11.3) For meta-learning, the multi-objective optimization algorithm NSGA-II[45] from the Optuna[64] package was leveraged with population size of 50, without mutation probability, with a crossover probability of 0.9, swapping probability of 0.5 and a fixed seed of 42.

The losses of labels and priors are weighed against each other by uncertainty weighing[44]. For this purpose, additional learnable parameters are introduced, that weigh the losses against each other. The loss is therefore, determined by: $\mathcal{L} = \sum_i (\frac{\mathcal{L}_{L,i}}{\sigma_{L,i}^2} + \log \sigma_{L,i}) + \sum_j (\frac{\mathcal{L}_{P,j}}{\sigma_{P,j}^2} + \log \sigma_{P,j})$ with labels i of set L and priors j of set P, and the learnable uncertainties associated with each label $\sigma_{L,i}$ and prior $\sigma_{P,j}$. For the *2.5D* DaRes, three 2D slices of the 3D images are fed in three channels of a 2D EfficientNet. The center slice of the cropped bounding box is used and two further peripheral slices, whose distance from the center slice is optimized by SEMPAI. For NN with branches, i.e., SEMPAI *Branches* and *AuxiliaryLosses&Branches*, a wrapper was built for the respective NN to introduce the priors in the fully connected layers.

For AutoML based on priors, i.e., SEMPAI *PriorsOnly*, the Tree-based Pipeline Optimization Tool (TPOT)[65] was employed. This algorithm combines identification of feature selection and suitable classifiers or regressors with Pareto optimization. We used 250 generations, a population size of 200, and grouping of the fibers. The combined train&dev set was forwarded to TPOT for training, and the internal cross-validation (CV) was set to two-fold to have a comparable data split ratio to the other components of SEMPAI. TPOT was restricted to methods with class probability output. The performance metric, e.g., *AUC* or $R^2$, of the internal CV is reported to SEMPAI and evaluated as a meta-loss, i.e., the model selected by SEMPAI can be a prior-only model based on AutoML.

The *total meta-loss* is a weighted sum of each label. The labels are weighted as a trade-off between sample size and importance of task, accordingly we set weights w = [*mdx*: 1.0, sepsis: 1.0, muscle type: 0.5, active force: 1.0, active force/pCa: 1.0, passive force: 1.0, pCa50 = 1.0]. In the trade-off between exploration and exploitation, multi-objective optimization algorithms are lending towards exploration as the performance for different tasks has to be optimized. Thus, the configuration space is sufficiently sampled although very unpromising regions of configuration space trials are still under-sampled. Selecting a criterion time for early termination of the trials is not trivial for multi-objective optimization trials. Therefore, a very non-conservative criterion was selected. Accordingly, SEMPAI does not compare trials for termination (and save computation time) as in more modern methods like Hyperband pruning[66]. The *total meta-loss* is smoothed by computation of the moving average of the last 10 epochs. A trial is terminated when the *total meta-loss* did not decrease for 50 subsequent epochs. The early stopping criterion is set active after the initial 75 epochs, resulting in at least 125 epochs performed per trial. The lowest meta-loss for each respective task is used to select the respective model for the task. For tasks with scarce data (pCa50, passive force), however, the *total meta-loss* is used for model selection.

To provide more insights about preferable individual configurations for each target label, we calculated SHapley Additive exPlanations (SHAP)[31] values. A random forest was trained to predict performance metrics of the dev set based on the configuration space. We utilize the Tree Explainer[46], which was explicitly designed for tree-based algorithms on this fitted model. We verify the stability of our results by fitting multiple forests with different random initializations and ensemble sizes (i.e., number of trees). Manually inspecting each resulting plot of two representative labels (*mdx* and active force/pCa) gave rise to the same interpretation.

*Rationale for standardization and configuration space:*
a. *Standardization:* Standardization is intended to minimize technical variance, which is usually present in biomarker research[67]. This technical variance can even lead to wrong conclusions of an AI system[68, 69]. To reduce the impact of technical variance, we slightly denoised the image and resampled to uniform isotropic voxel size. Cropping reduces the dimensionality of the images and DL can focus only on relevant regions. The alignment of fibers via registration helps to minimize the bounding box and can increase the convergence of the learning process, because convolutional NNs such as the employed EfficientNet are not rotation invariant.
b. *Configuration space:*
   - The benefit of contrast enhancement for visual recognizability of structures is undisputed. However, it is not yet understood if this enhancement adds value for training an AI. Therefore, SEMPAI validates this explicitly and exemplarily for the MCLAHE[38] algorithm.
   - Random erasing[48] regularizes the learning process by enforcing the use of multiple image regions for inference, theoretically resulting in a more robust prediction. In the case of localized biologically-relevant image properties, however, deleting this location naturally leads to a mis-evaluation of the image and a decrease in predictive performance. We thus use random sampling as a measure for the importance of localized image features.
   - Down-sampling and multi-view representations may support learning by minimizing overfitting. It is scientifically interesting to understand the importance of resolution for learning phenotypes and function, since microscopy research targets finer resolution (lower pixel size), often at the expense of reduced throughput. We interpret SEMPAI's decision w.r.t. down-sampling to elaborate how important the image resolution is for a given learning task. In analogy, we evaluate whether to use 3D data directly for learning, or to draw representative 2D slices. We test whether using lower dimensional DaRe as NN input via down-sampling (reduced voxel size) and sub-sampling (2.5D vs 3D), improves convergence. The benefit of dimensionality reduction in DL is controversial[70, 71]. Choosing the spacing of the representative slices is also of biological interest. It allows interpretation of where relevant information is located in 3D, i.e., by interpretably sub-sampling a lower-dimensional DaRe from a higher dimensional volume. By this, the origin of the biological information can be narrowed down.
   - To test the importance of priors, we use several prior integration methods. Besides both extremes, *NoPriors* and the *PriorsOnly*, we use priors as auxiliary tasks, as branches or as a combination of the latter two, to define a scale of prior integration from "weak to strong".

By defining the priors as auxiliary tasks, they are predicted simultaneously to the labels. Thus, the image filters evolve to predict these auxiliary tasks as well. Since the prior is only indirectly available for learning a label, we consider it as weak prior integration. With the branches approach, the priors are passed on directly to the fully connected layers of the NN, i.e., theoretically, the NN can completely dispense with the additional image information, which is why we consider it as strong prior integration. Multi-branch approaches have recently been shown to have positive convergence properties[72] for learning. We define the combination of both methods as an even stronger prior integration. Finally, the use of priors with AutoML, i.e., without images and DL, is defined as the "maximum" of our prior integration scale. Such feature-based ML approaches can occasionally outperform DL[73]. In the optimization of SEMPAI, the added value of the priors for the learning process is evaluated. If models with the hypothesis-driven priors are superior to models without, or if a prior-only model shows the same performance as the best DL model, the hypothesis that the prior describes the state of the label well can be considered true. The researcher can thus test hypotheses and these are verified by SEMPAI and, in the case of models with DL, also refined. The biological information of the prior knowledge is evaluated.

- Further adaptations: NN-specific parameters are more technical and less interpretable, but need to be adapted to prior integration and DaRe at hand to achieve a global optimum. The NN capacity is adjusted, as it must be adapted to the available amount of data and the complexity of the learning task. Also, further NN properties like batch size, learning rate, momentum, and optimizer have to be fine-tuned. Gradient clipping, i.e., restricting the gradients, has been theoretically shown to accelerate convergence[74] and its benefit is evaluated. Also, the sampling of the data can be modified by imbalance sampling. Our employed augmentation uses rotations, shifts, and additive noise patterns, which were identified as variations in the data after inspection of the images by domain experts. Thus, this step can also be interpreted as prior knowledge integration. Augmentation introduces invariance towards the applied modifications to the learning process.

*Computation details:* SEMPAI computed 15 days on a workstation equipped with NVMe SSD, Nvidia RTX3090 GPU and Intel Core-i9 10850k CPU (10 cores of 3.6 GHz), resulting in a total of 1,500 evaluated trials. To decrease the computational cost for evaluating 2D configurations, the slices were loaded by reading parts of the memory-mapped 3D volume. For (3D) augmentations, some operations employ TorchIO[75], and where possible, augmentations were computed on the GPU. Automated mixed precision (AMP) of PyTorch was used in addition to multiple workers and pinned memory. To be able to use sufficiently large batches for 3D data, SEMPAI utilizes gradient accumulation.

# SEMPAI: a Self-Enhancing Multi-Photon Artificial Intelligence for prior-informed assessment of muscle function and pathology

Alexander Mühlberg*, Paul Ritter, Simon Langer, Chloë Goossens, Stefanie Nübler, Dominik Schneidereit, Oliver Taubmann, Felix Denzinger, Dominik Nörenberg, Michael Haug, Wolfgang H. Goldmann, Andreas K. Maier, Oliver Friedrich, Lucas Kreiss

**Supporting Information 1: High-resolution examples for IQ of MPM images after cross-study standardization.**

The data is accessible at: https://sempai-mbt.github.io/

**1 Study A – SEPSIS**
Title/Caption: High-resolution PDF of exemplary fibers of Study A (Sepsis) after standardization without contrast-enhancement.
**2 Study A – CE SEPSIS**
Title/Caption: High-resolution PDF of exemplary fibers of Study A (Sepsis) after standardization with contrast-enhancement.
**3 Study B1 – MDX & ACTIVEFORCE**
Title/Caption: High-resolution PDF of exemplary fibers of Study B1 (mdx, active force) after standardization without contrast-enhancement.
**4 Study B1 – CE MDX & ACTIVEFORCE**
Title/Caption: High-resolution PDF of exemplary fibers of Study B1 (mdx, active force) after standardization with contrast-enhancement.
**5 Study B2 – MDX & PCA50**
Title/Caption: High-resolution PDF of exemplary fibers of Study B2 (mdx, pCa50) after standardization without contrast-enhancement.
**6 Study B2 – CE MDX & PCA50**
Title/Caption: High-resolution PDF of exemplary fibers of Study B2 (mdx, pCa50) after standardization with contrast-enhancement.
**7 Study C – MDX & PASSIVEFORCE**
Title/Caption: High-resolution PDF of exemplary fibers of Study C (mdx, passive force) after standardization without contrast-enhancement.
**8 Study C – CE MDX & PASSIVEFORCE**
Title/Caption: High-resolution PDF of exemplary fibers of Study C (mdx, passive force) after standardization with contrast-enhancement.
**9 Study D – MDX & MUSCLETYPE**
Title/Caption: High-resolution PDF of exemplary fibers of Study D (mdx, muscle type) after standardization without contrast-enhancement.
**10 Study D – CE MDX & MUSCLETYPE**
Title/Caption: High-resolution PDF of exemplary fibers of Study D (mdx, muscle type) after standardization with contrast-enhancement.

**Supporting Information 2: Examples for sample-level explanations for image and priors**

**Configuration of trial with lowest total meta-loss**
*SEMPAI.config = {'augmentation': True,*
*'batch size': 'small',*
*'clipping': True,*
*'complexity': 2,*
*'cyclic learning rate': False,*
*'dimension': '2D_5',*
*'downsampling': True,*
*'enhancement': True,*
*'imbalanced dataset sampling': False,*
*'lr': 0.0540359659501924,*
*'momentum': 0.9134831480715462,*
*'optimizer': 'SGD',*
*'prior': "aux_loss&branches",*
*'random erasing': True}*

**Task *mdx*:** Example #1

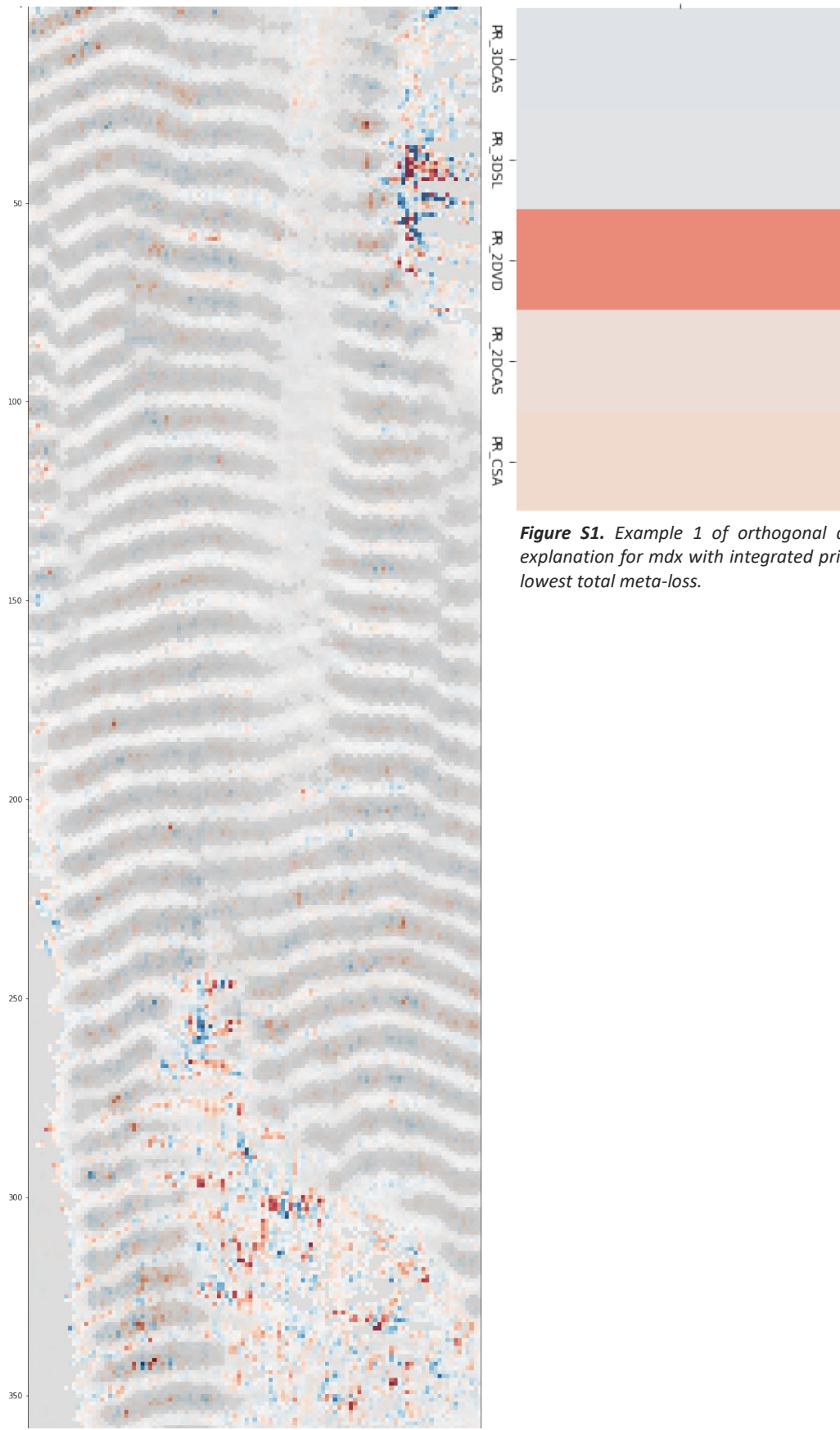

*Figure S1. Example 1 of orthogonal decision explanation for mdx with integrated priors and lowest total meta-loss.*

Example #2

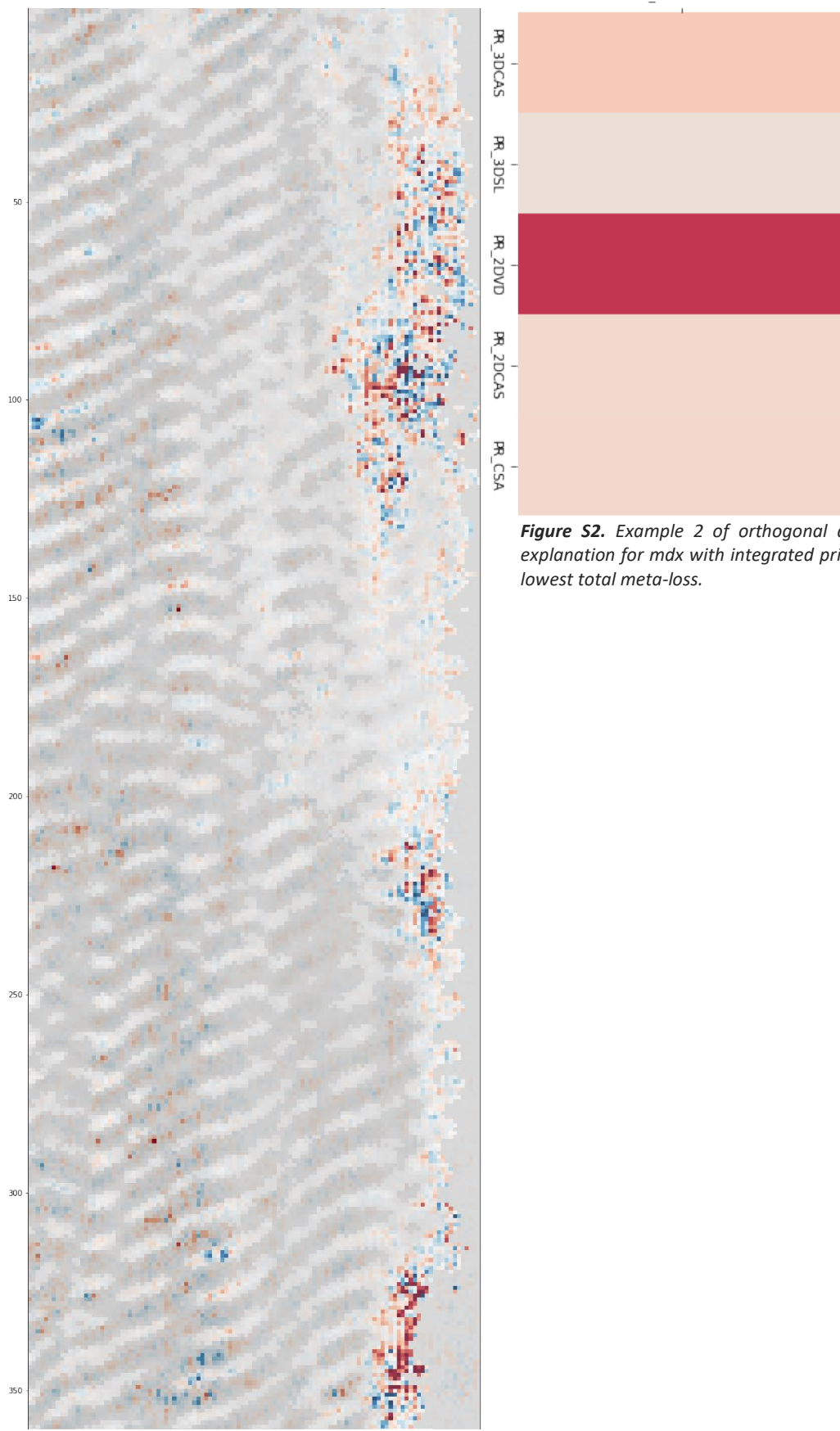

*Figure S2.* Example 2 of orthogonal decision explanation for mdx with integrated priors and lowest total meta-loss.

**Task Sepsis:** Example #1

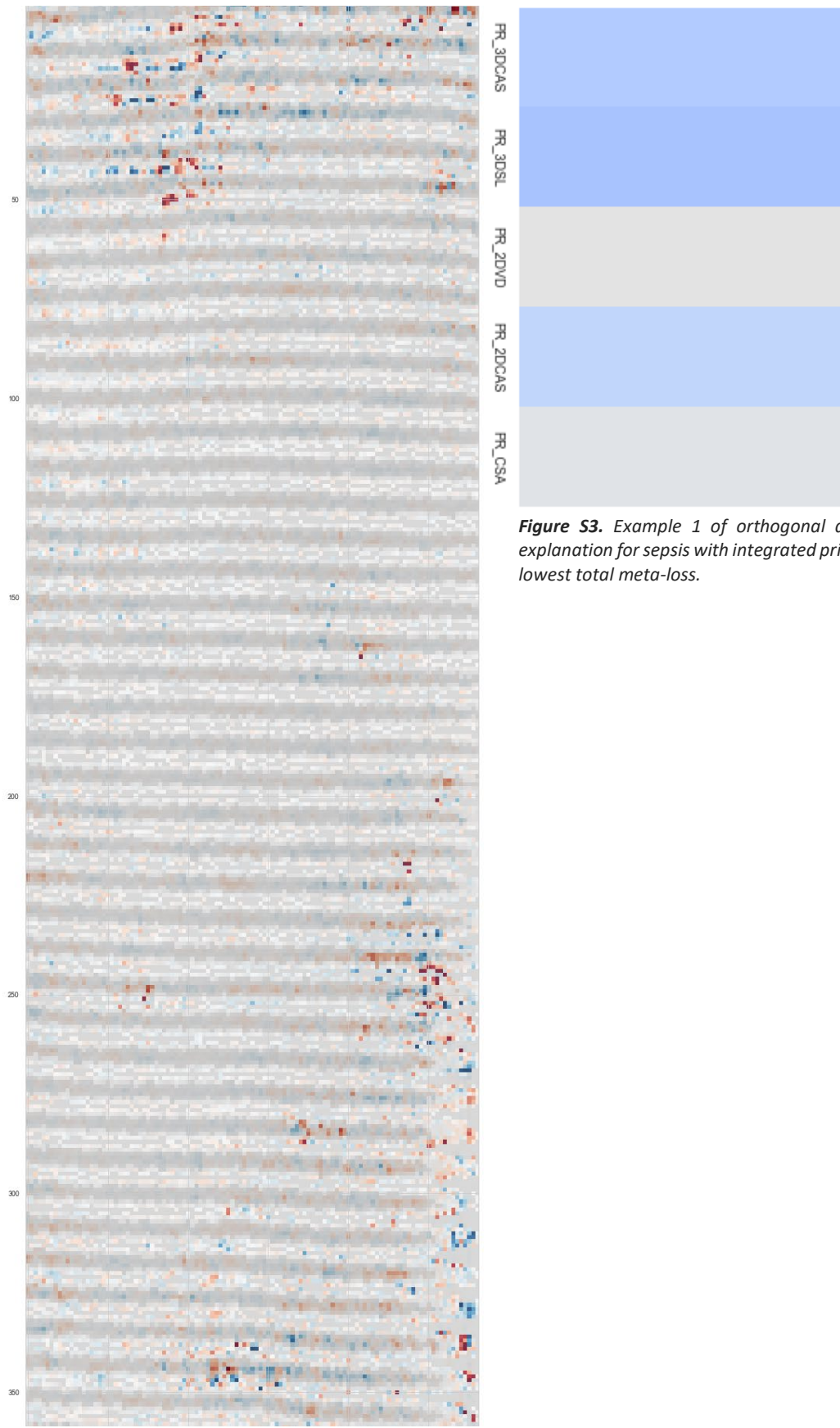

*Figure S3.* Example 1 of orthogonal decision explanation for sepsis with integrated priors and lowest total meta-loss.

Example #2

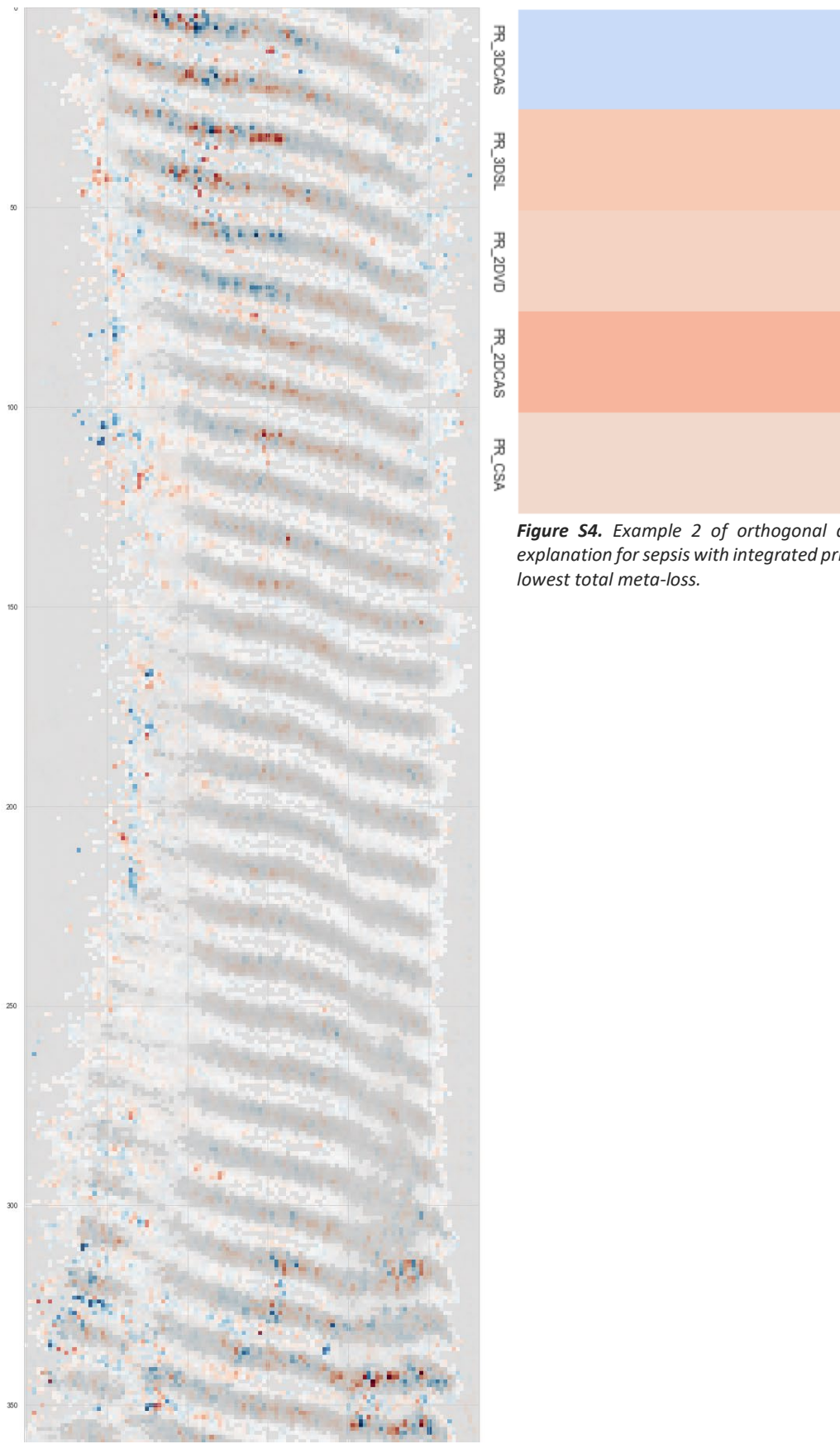

*Figure S4.* Example 2 of orthogonal decision explanation for sepsis with integrated priors and lowest total meta-loss.

**Task Muscle Type:** Example #1

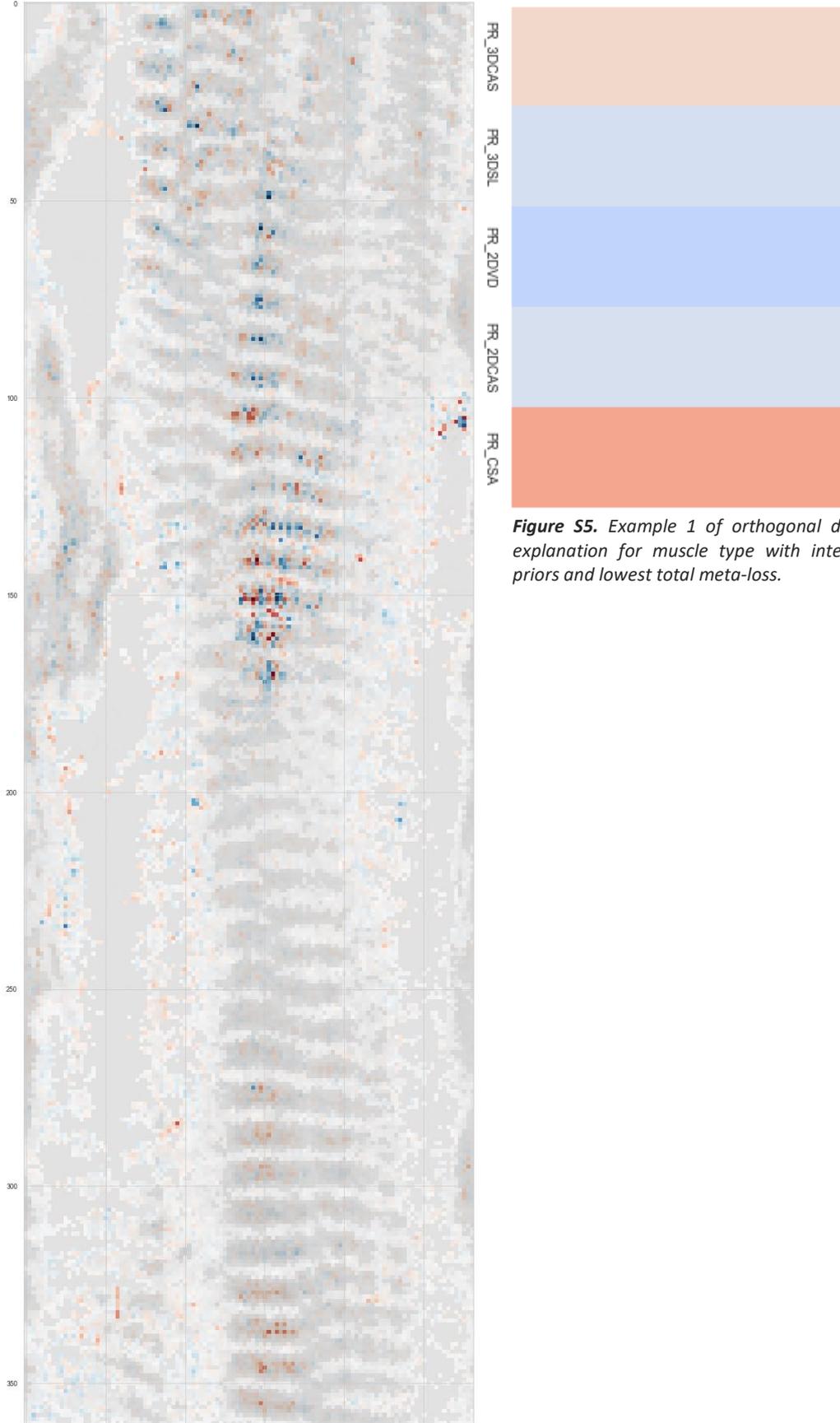

*Figure S5.* Example 1 of orthogonal decision explanation for muscle type with integrated priors and lowest total meta-loss.

Example #2

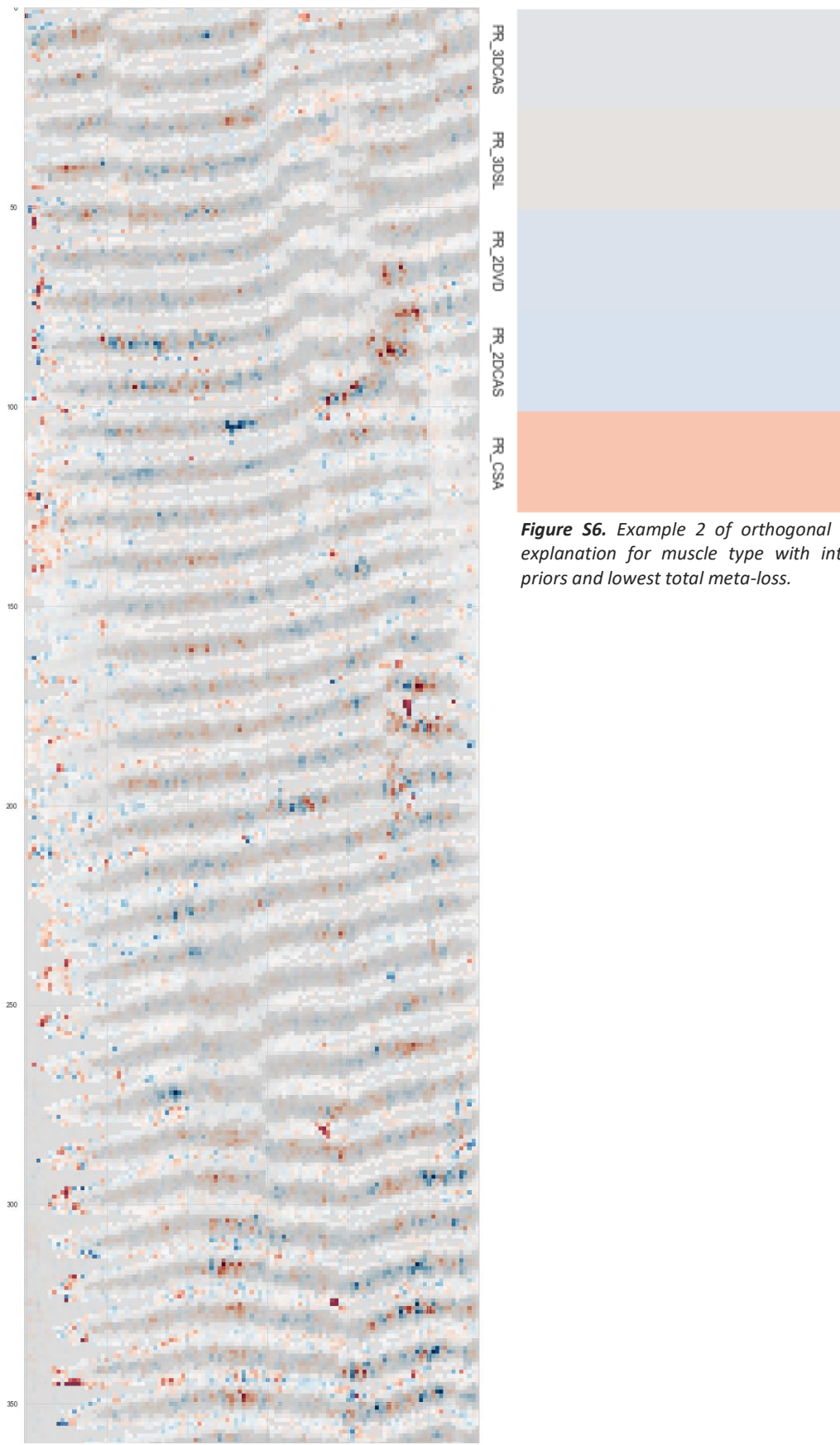

***Figure S6.*** *Example 2 of orthogonal decision explanation for muscle type with integrated priors and lowest total meta-loss.*

**Task Active Force:** Example #1

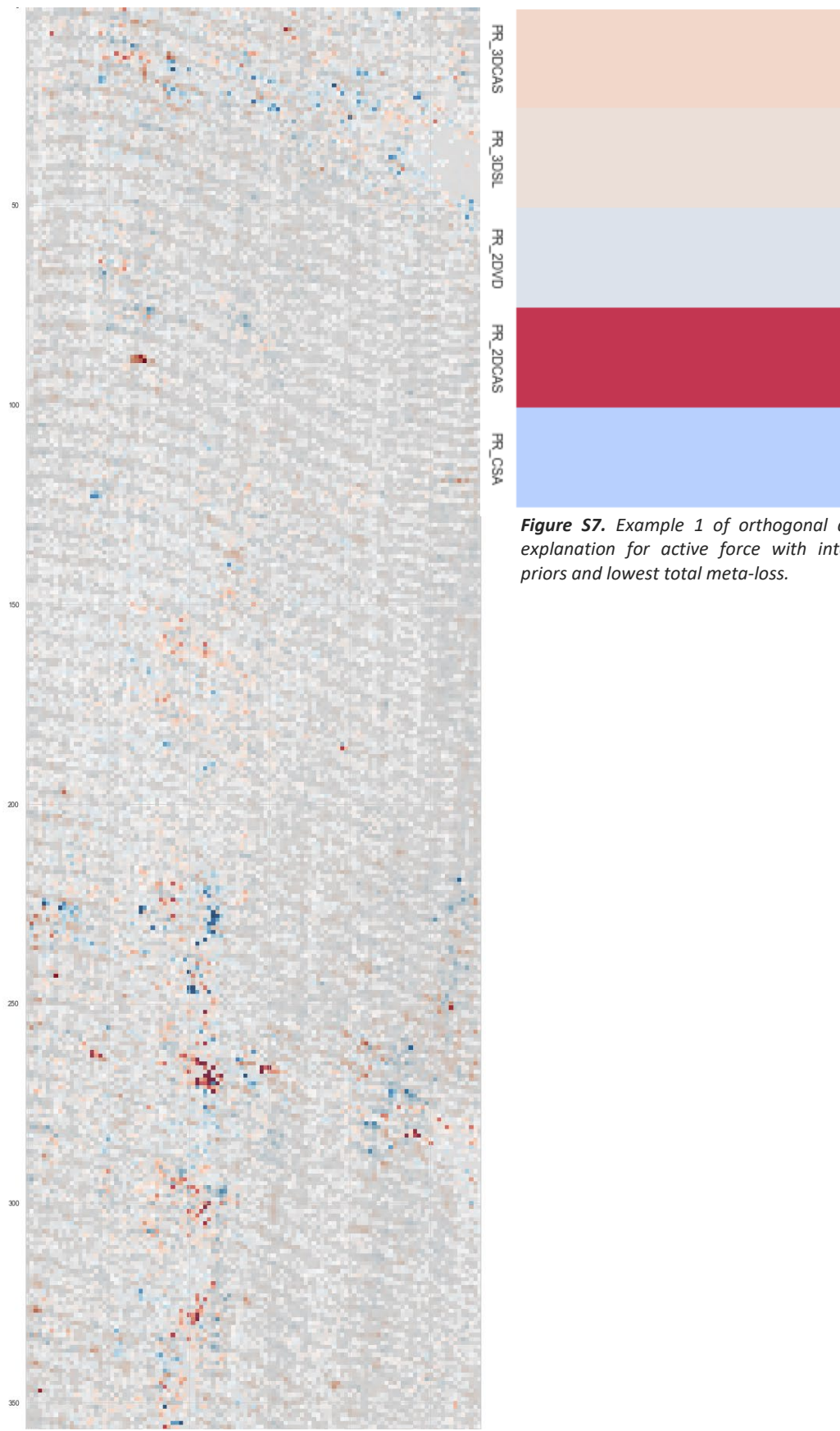

*Figure S7.* Example 1 of orthogonal decision explanation for active force with integrated priors and lowest total meta-loss.

Example #2

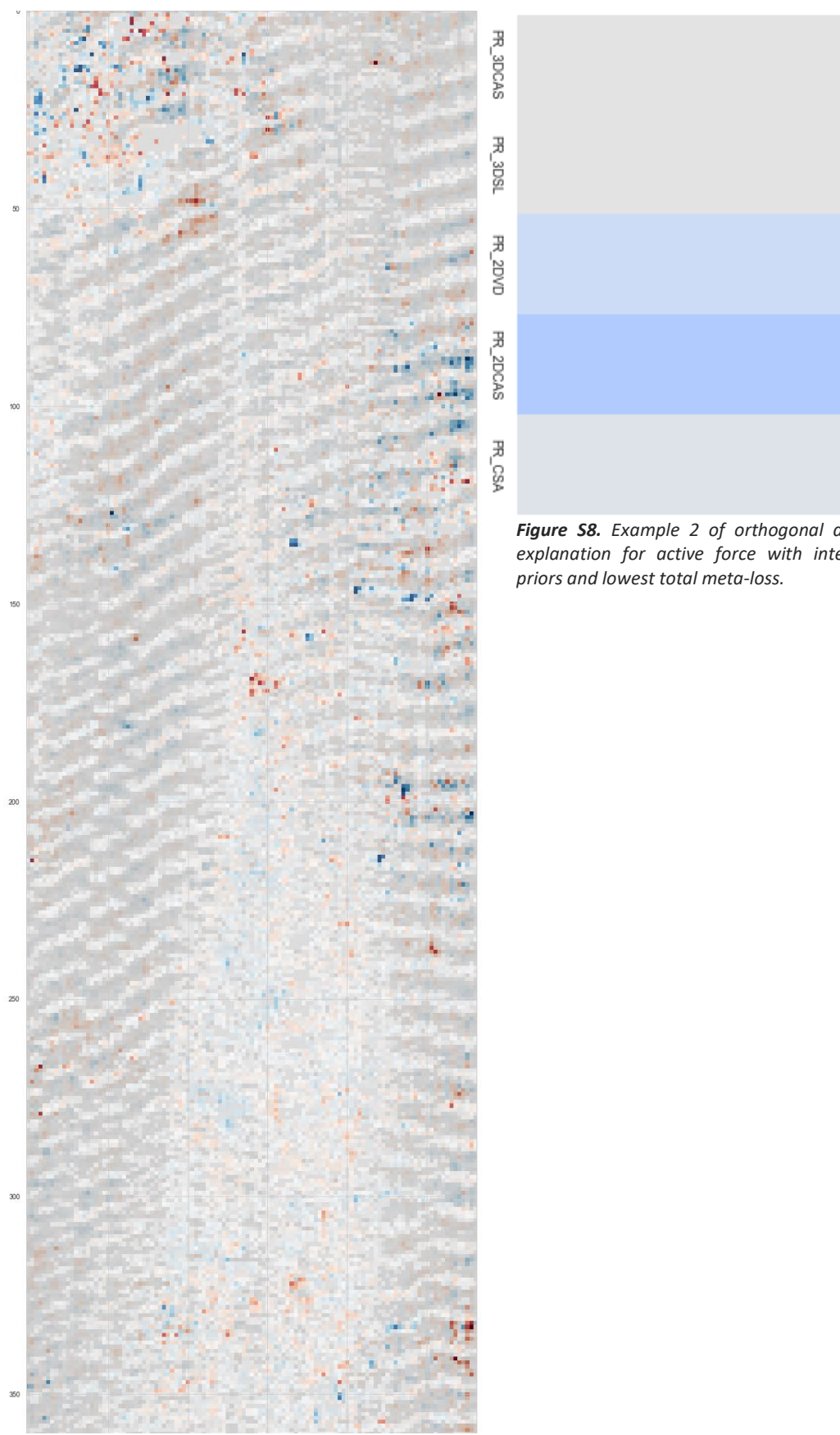

*Figure S8.* Example 2 of orthogonal decision explanation for active force with integrated priors and lowest total meta-loss.

**Task Active Force/pCa:** Example #1

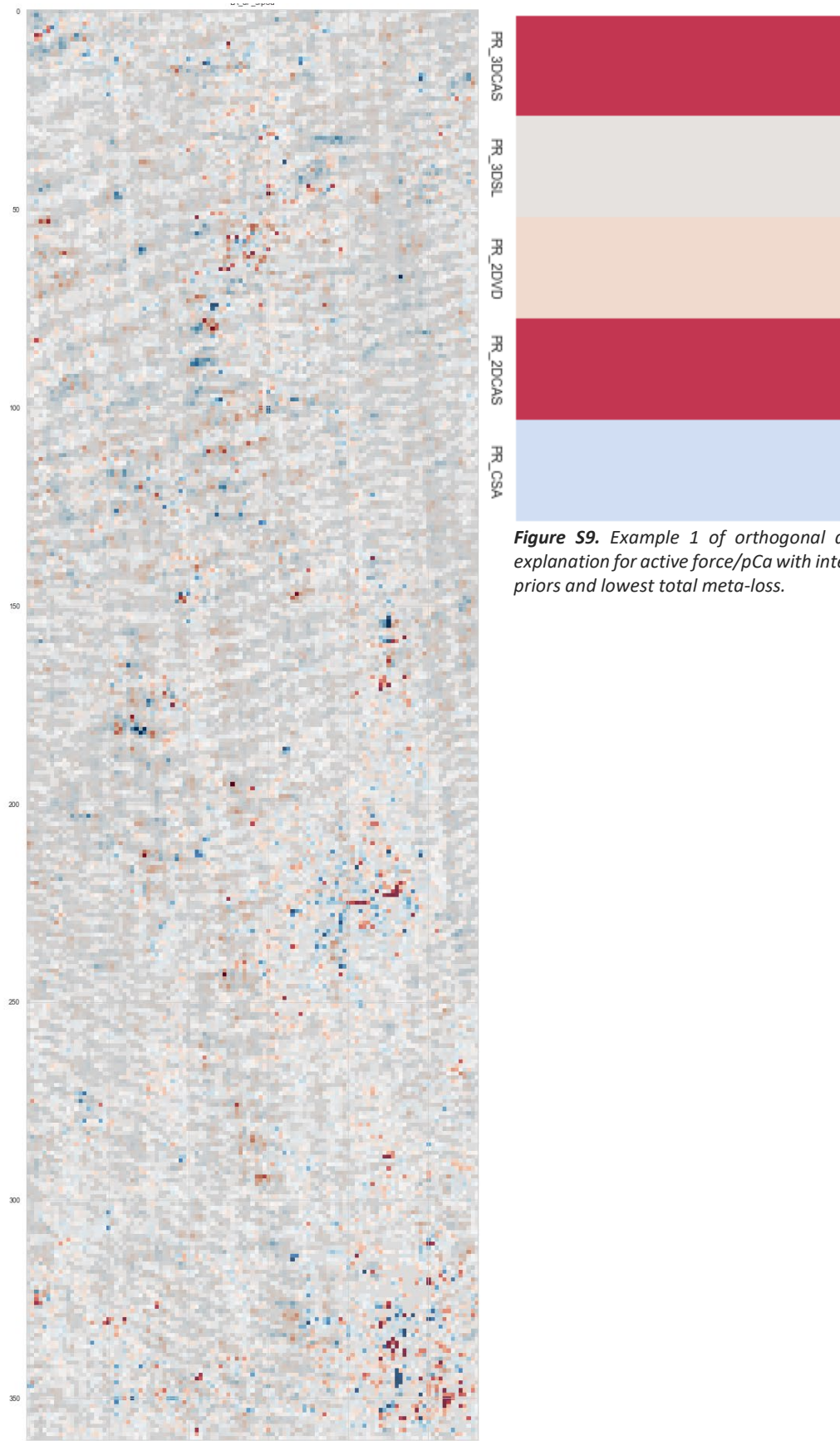

*Figure S9.* Example 1 of orthogonal decision explanation for active force/pCa with integrated priors and lowest total meta-loss.

Example #2

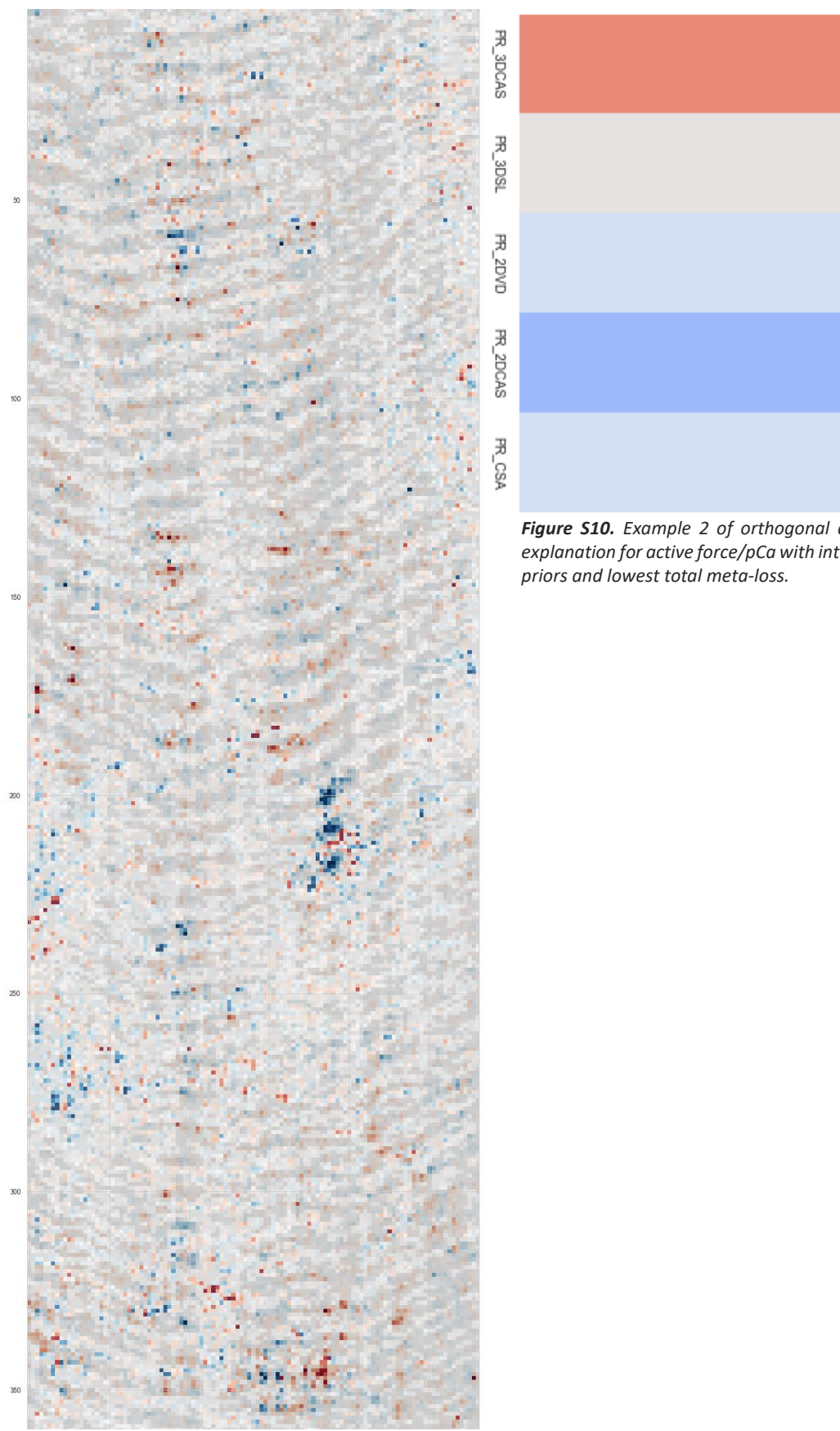

*Figure S10.* Example 2 of orthogonal decision explanation for active force/pCa with integrated priors and lowest total meta-loss.

**Task pCa50:** Example #1

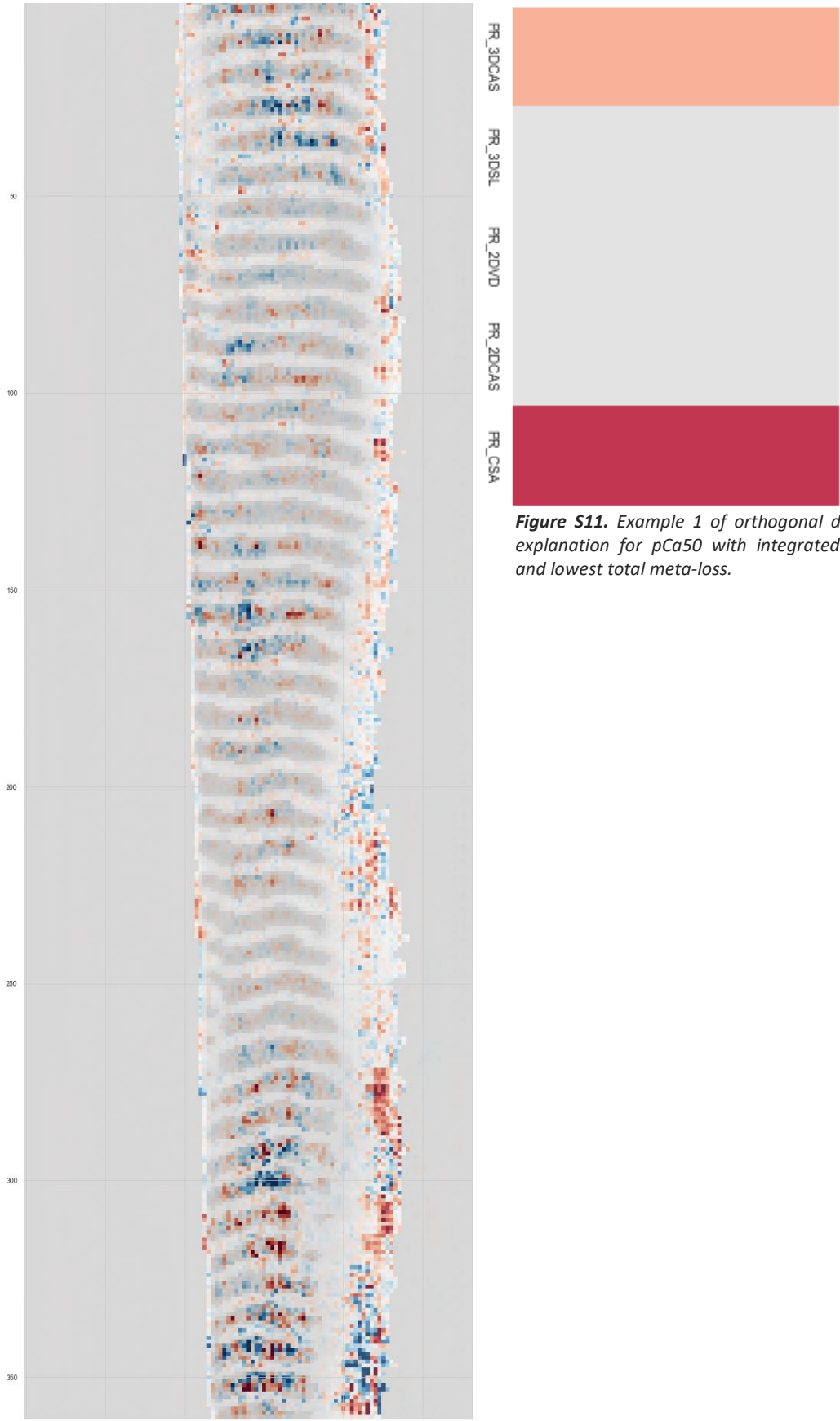

*Figure S11.* Example 1 of orthogonal decision explanation for pCa50 with integrated priors and lowest total meta-loss.

Example #2

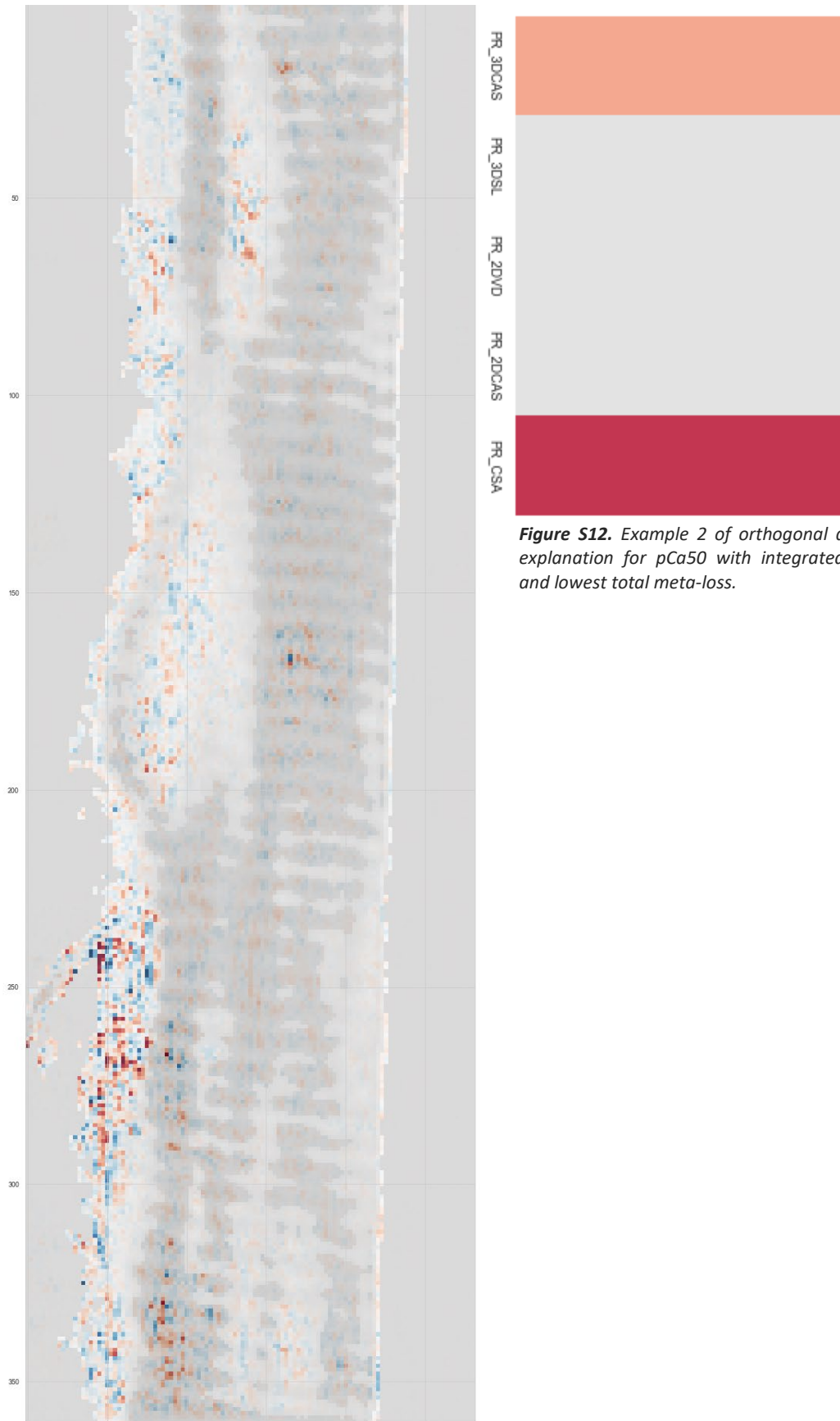

***Figure S12.*** Example 2 of orthogonal decision explanation for pCa50 with integrated priors and lowest total meta-loss.